\documentclass[dvipsnames,format=sigconf,nonacm]{acmart}

\setcopyright{acmlicensed}
\copyrightyear{2018}
\acmYear{2018}
\acmDOI{XXXXXXX.XXXXXXX}

\acmJournal{JDS}
\acmVolume{37}
\acmNumber{4}
\acmArticle{111}
\acmMonth{8}

\usepackage{graphicx}     
\usepackage{subcaption}   
\usepackage{amsmath}
\usepackage{algorithm}
\usepackage{algpseudocode}

\begin{document}

\title{ROIDS: Robust Outlier-Aware Informed Down-Sampling}

\author{Alina Geiger}
\affiliation{
  \institution{Johannes Gutenberg University}
  \city{Mainz}
  \country{Germany}}
\email{geiger@uni-mainz.de}

\author{Martin Briesch}
\affiliation{
  \institution{Johannes Gutenberg University}
  \city{Mainz}
  \country{Germany}}
\email{briesch@uni-mainz.de}

\author{Dominik Sobania}
\affiliation{
  \institution{University of Duisburg-Essen}
  \city{Essen}
  \country{Germany}}
\email{dominik.sobania@uni-due.de}

\author{Franz Rothlauf}
\affiliation{
  \institution{Johannes Gutenberg University}
  \city{Mainz}
  \country{Germany}}
\email{rothlauf@uni-mainz.de}

\renewcommand{\shortauthors}{Geiger et al.}

\acmArticleType{Review}

\keywords{Symbolic regression, genetic programming, parent selection, down-sampling}

\begin{abstract}
Informed down‑sampling (IDS) is known to improve performance in symbolic regression when combined with various selection strategies, especially tournament selection. However, recent work found that IDS's gains are not consistent across all problems. Our analysis reveals that IDS performance is worse for problems containing outliers. IDS systematically favors including outliers in subsets which pushes GP towards finding solutions that overfit to outliers.  
To address this, we introduce ROIDS (Robust Outlier‑Aware Informed Down‑Sampling), which excludes potential outliers from the sampling process of IDS. With ROIDS it is possible to keep the advantages of IDS without overfitting to outliers and to compete on a wide range of benchmark problems. This is also reflected in our experiments in which ROIDS shows the desired behavior on all studied benchmark problems. 
ROIDS consistently outperforms IDS on synthetic problems with added outliers as well as on a wide range of complex real-world problems, surpassing IDS on over 80\% of the real-world benchmark problems. Moreover, compared to all studied baseline approaches, ROIDS achieves the best average rank across all tested benchmark problems. This robust behavior makes ROIDS a reliable down-sampling method for selection in symbolic regression, especially when outliers may be included in the data set.
\end{abstract}

\maketitle

\section{Introduction}

Down-sampled selection strategies have been found to enhance the performance and generalization ability of Genetic Programming (GP) in areas like program synthesis~\cite{Hernandez.2019, Ferguson.2020} and symbolic regression~\cite{geiger.2023, geiger2024comprehensive}, while simultaneously reducing the computational costs~\cite{Goncalves.2012, geigertournament}. Instead of evaluating the candidate solutions on all available training cases in each selection event, down-sampled selection methods only use a subset of the training cases to evaluate the quality of the candidate solutions in a population. A na\"ive approach is random down-sampling (RDS)~\cite{Goncalves.2012}, where in each generation a subset of training cases is randomly sampled.

Unfortunately, subsets of training cases generated by RDS often contain many similar cases that bring in only little new information, while it potentially excludes other, more rare edge or corner cases which are often much more informative for GP. 
Ignoring such rare but relevant edge cases for several generations potentially harms the overall performance of RDS \cite{boldi2024informed}. 

Therefore, informed down-sampling (IDS)~\cite{boldi2024informed} has been proposed to create diverse subsets, where each case is intended to test different niches. 
To reach this goal, IDS evaluates a fraction of the population on all cases in the training set and then constructs for each case a vector which measures the performance of the evaluated candidate solutions for this case. As next step, it computes the distance matrix between all vectors and selects the subset of the population that maximizes the pairwise distances in the subset. 
Based on this procedure, IDS selects a diverse subset of training cases with regards to the behavior of the current population. 

Originally, IDS was described for program synthesis problems, where GP using IDS is able to find better results~\cite{boldi2024informed}. It also increases performance in the domain of symbolic regression, although the performance gains are not consistent over the range of different problems. On some real-world regression problems, IDS performs poorly and even worse than RDS~\cite{geiger2025performance, geigertournament}. 

To gain a deeper understanding of the differences in performance, we analyze the behavior of IDS for regression problems in detail. We find that IDS outperforms RDS for problems with rare cases but performs worse when outliers are present, which is often the case in real-world regression problems. To overcome this problem, we propose \textbf{R}obust \textbf{O}utlier-Aware \textbf{I}nformed \textbf{D}own-\textbf{S}ampling (\textbf{ROIDS}) which excludes potential outliers from the sampling process of IDS. With ROIDS, we are able to leverage the advantages of IDS without overfitting to outliers to compete even on real-world regression problems. 

We compare the performance of ROIDS, IDS, and RDS on a set of synthetic problems with and without outliers as well as for a representative set of real-world regression problems. We find that ROIDS performs significantly better than IDS on all synthetic problems in the presence of outliers and never worse. Moreover, the average ranking of ROIDS across all synthetic problem is better compared to IDS and RDS. 

Also for real-world problems, we find that ROIDS outperforms IDS on problems where IDS only produces poor results. Furthermore, we find that ROIDS outperforms IDS even when we apply an outlier removal technique beforehand. We visualize the problems using UMAPs (Uniform Manifold Approximation and Projection~\cite{mcinnes2018umap}) for dimensionality reduction and observe that ROIDS, as intended, focuses less on potential outliers. Overall, ROIDS achieves a lower error than IDS on over 80\% of the studied problems. Again, the average ranking of 1.7 achieved by ROIDS is better compared to the rankings of RDS and IDS, which are 2.5 and 3, respectively. This indicates that ROIDS performs consistently well across different problems. 

From a computational perspective, ROIDS comes at almost no additional cost, but leads to similar or better search performance than IDS, especially for datasets with outliers. As the implementation of ROIDS is of low complexity and it always achieves a better or comparable performance, we recommend researches as well as practitioners to replace IDS with ROIDS for symbolic regression. 

Following this introduction, we present the related work in Sect.~\ref{sec:related_work} and an analysis of the behavior of IDS in Sect.~\ref{sec:analysis}. In Sect.~\ref{sec:method}, we describe ROIDS.  Section~\ref{sec:experimental_setup} describes our experimental setup, followed by a presentation of our results in Sect.~\ref{sec:results}. The conclusions are drawn in Sect.~\ref{sec:conclusions}.

\section{Related work}\label{sec:related_work}

Down-sampling strategies evaluate candidate solutions using only a subset of the training cases in each generation of an evolutionary run. This reduces the computational cost per generation, which allows the reallocation of the saved evaluations to either a longer search or larger population sizes~\cite{Helmuth.2020c,briesch2023trade}.  

A simple yet effective down-sampling strategy is RDS, which randomly samples a fraction of the training cases in each generation. Early work found, that RDS reduces overfitting as well as bloat in GP runs for symbolic regression problems~\cite{Goncalves.2012}. 
More recently, RDS combined with lexicase selection~\cite{spector.2012, Helmuth.2014} increased the problem-solving success in the domain of program synthesis~\cite{Hernandez.2019, Ferguson.2020, Schweim.2022} and symbolic regression~\cite{geiger.2023, geiger2024comprehensive}. The main advantage of down-sampled lexicase selection is its ability to evaluate a higher number of candidate solutions with the same computational costs~\cite{Helmuth.2021, Helmuth.2020c}. 

However, as RDS selects training cases randomly, it does not prefer edge or corner cases which would be relevant for GP. Instead, such cases may be excluded from consideration for several generations, which means that candidate solutions are not be tested for certain important behaviors. Therefore, \citet{boldi2024informed} proposed IDS to create more diverse subsets of training cases. In detail, every $l$ generations, IDS evaluates the performance of a sample of the population using all training cases to identify the cases which are solved by different candidates. If cases are solved by different candidates, these cases more likely test distinct behaviors and therefore 
contribute to a more diverse subset of training cases. The literature confirms these design decisions, as recent work finds that IDS outperforms RDS in the domain of program synthesis~\cite{boldi2024informed, Boldi2023.static} and has been studied in combination with different selection methods~\cite{Boldi2023.problemsolvingbenefits, boldi2024untangling}. 

Moreover, IDS also improves the performance of tournament selection in the domain of symbolic regression~\cite{geiger2025performance}. In combination with IDS, tournament selection performs and behaves similar to lexicase selection but with the benefit of being significantly faster than the lexicase variant~\cite{geigertournament}. 

However, recent work by ~\citet{geiger2025performance, geigertournament} observed that for some regression problems IDS performs poorly and even worse than RDS. The reason for this behavior is still unclear.

\section{The Limits of IDS in the Presence of Outliers}\label{sec:analysis}

In this section, we illustrate the behavior of IDS for several variants of the $2$D \texttt{nguyen-6} problem~\cite{uy2011semantically}, as this allows us to demonstrate why and how the existence of outliers in a dataset negatively affects the performance of IDS. This section is for illustrative purposes only, for more details and results, we refer to Sects.~\ref{sec:experimental_setup} and \ref{sec:results}.

IDS aims at creating a subset of relevant and important training cases. It tries to avoid considering those cases which bring in only little and redundant information about the problem at hand~\cite{boldi2024informed}.
For program synthesis problems, \citet{boldi2024informed} found that IDS includes highly informative edge cases more frequently. 
This is advantageous for program synthesis problems which are uncompromising in nature and noise free, meaning a solution needs to solve all training and test cases perfectly.
However, in symbolic regression problems the data might be noisy and inhibit outliers. Fitting a solution to this noise and outliers exactly is not desirable~\cite{lopez2017ransac} and including these cases in a subset might have negative effects.

\begin{figure}
    \centering
    \begin{subfigure}[b]{0.235\textwidth}
        \centering
        \includegraphics[width=\linewidth]{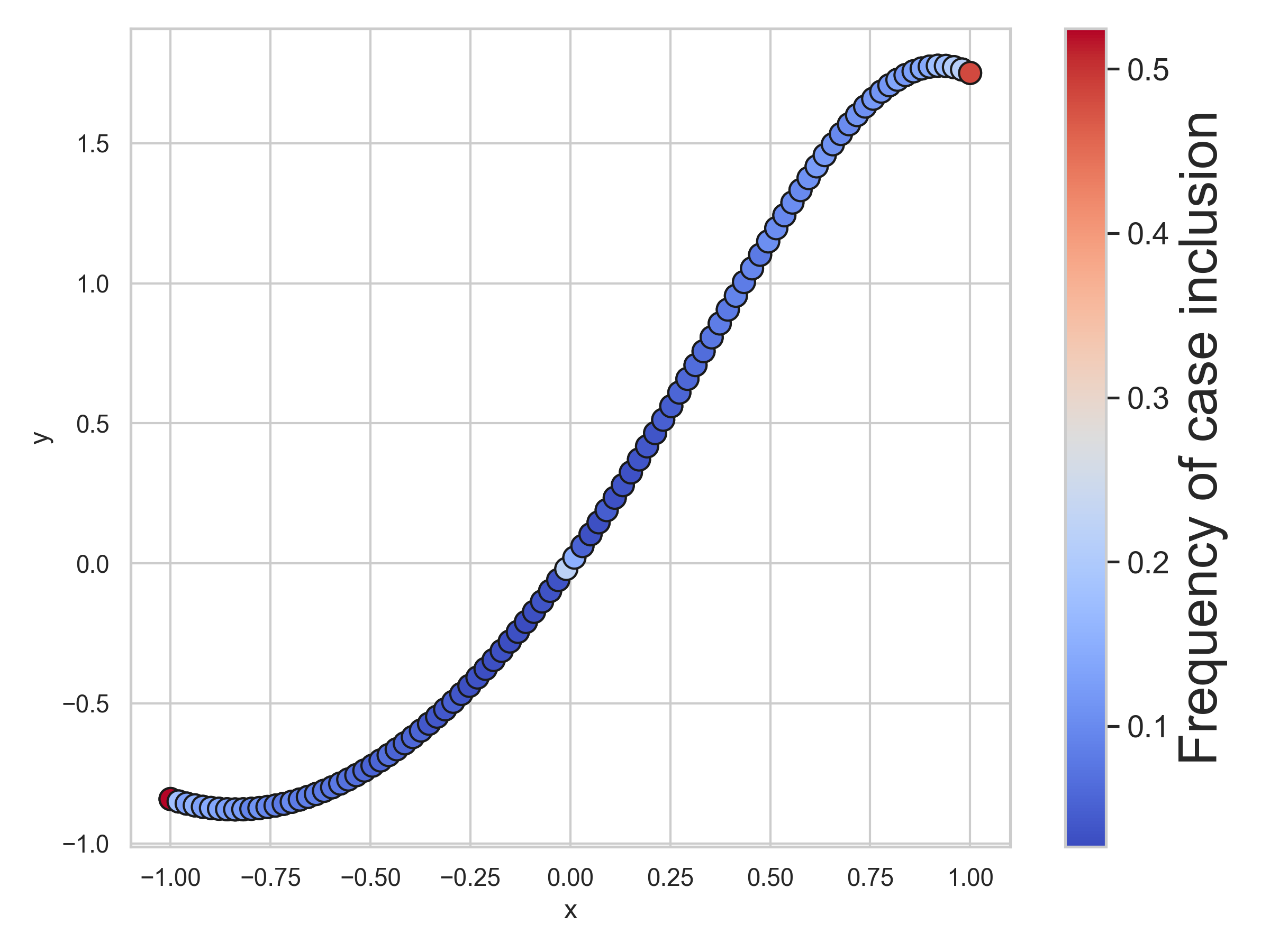}
\caption{Even distribution of \\training cases.}
        \label{subfig:even_ids}
    \end{subfigure}
    \hfill
    \begin{subfigure}[b]{0.235\textwidth}
        \centering
        \includegraphics[width=\linewidth]{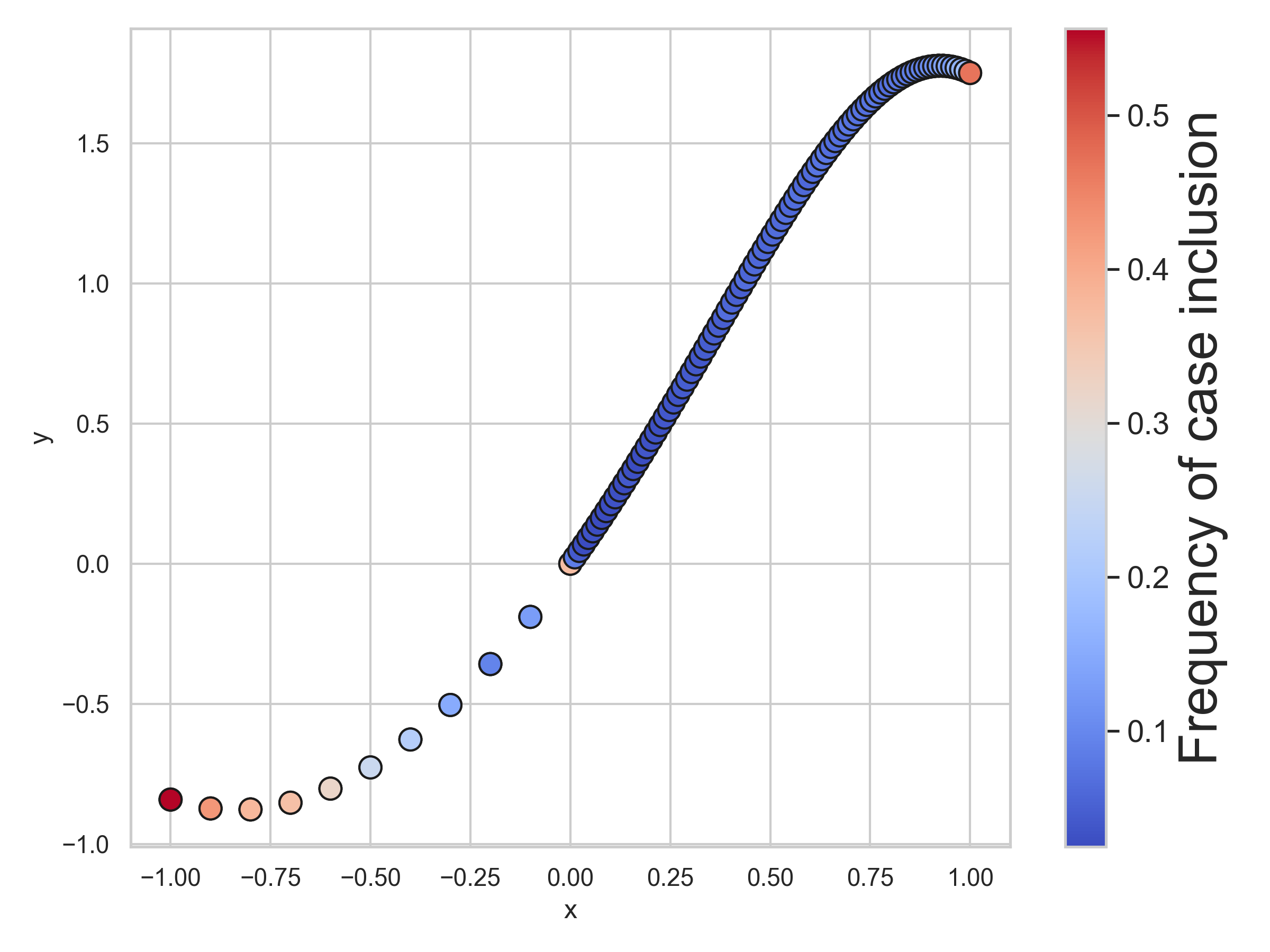}
        \caption{Uneven distribution of \\training cases.}
        \label{subfig:uneven_ids}
    \end{subfigure}
    \begin{subfigure}[b]{0.235\textwidth}
        \centering
        \includegraphics[width=\linewidth]{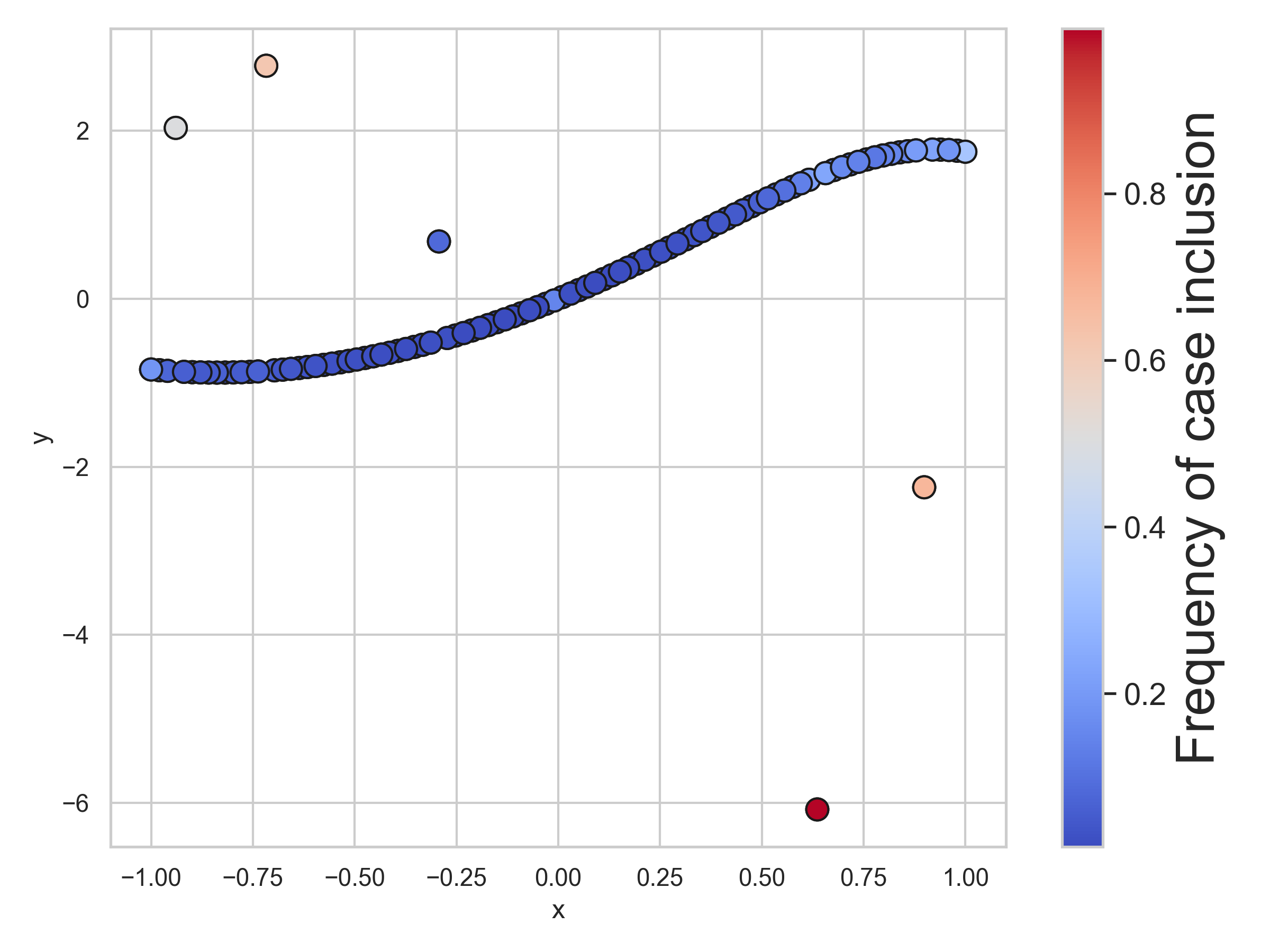}
        \caption{Even distribution of \\training cases and 5\% outliers.}
        \label{subfig:even_outliers_ids}
    \end{subfigure}
    \hfill
    \begin{subfigure}[b]{0.235\textwidth}
        \centering
        \includegraphics[width=\linewidth]{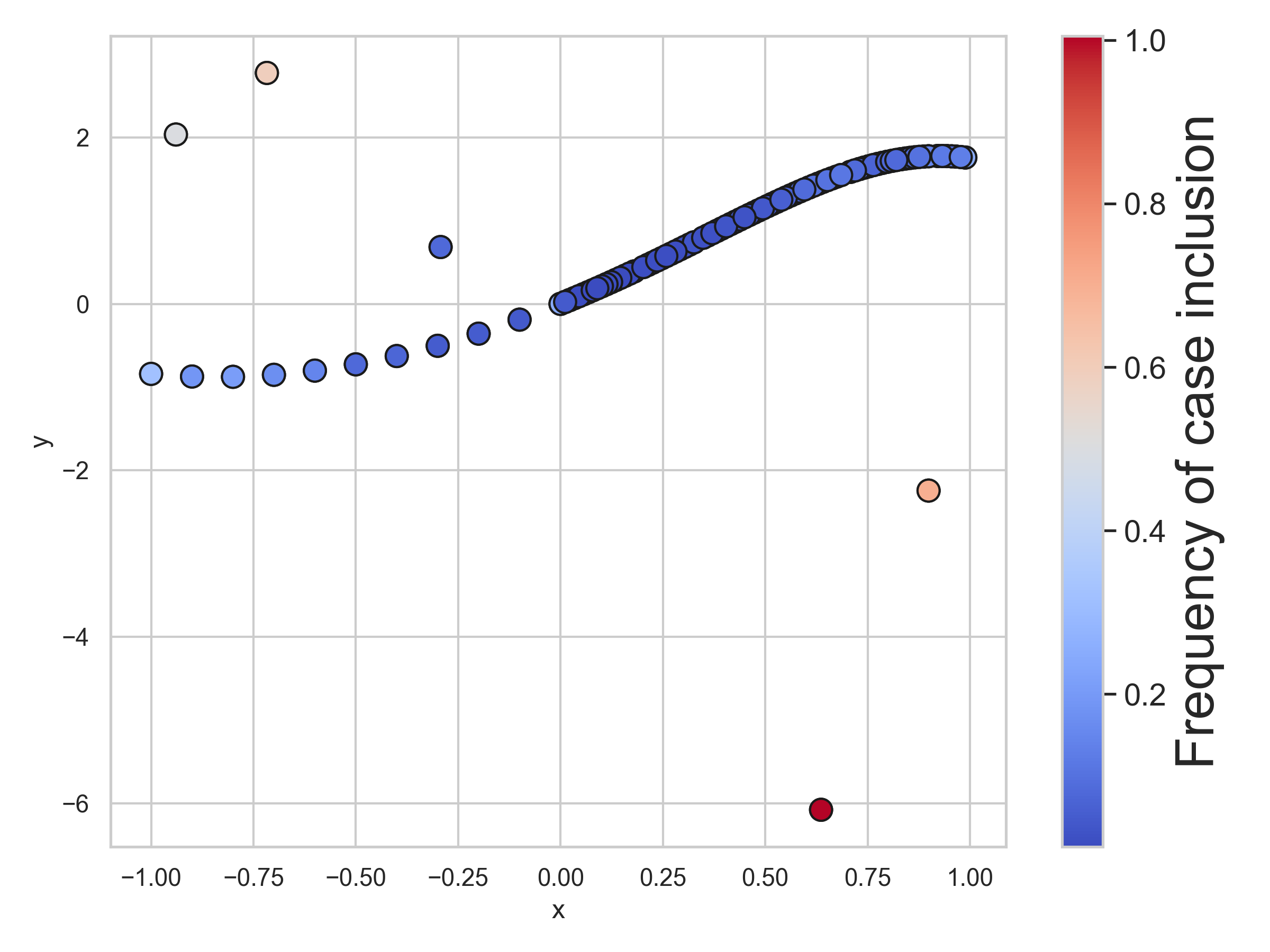}
        \caption{Uneven distribution of \\training cases and 5\% outliers.}
        \label{subfig:uneven_outliers_ids} 
    \end{subfigure}
    \caption{
    Color-coded frequencies of including a training case in the subsets when using IDS. Results are for different variants of the \texttt{nguyen-6} problem. The top row contains two outlier-free variants of the problem, whereas the bottom row includes 5\% outliers.}
    
    \label{fig:frequency_ids}
\end{figure}

Therefore, we exemplarily analyze the behavior of IDS for the 2D \texttt{nguyen-6} regression problem~\cite{uy2011semantically} with different distributions of training cases and number of outliers (for details of the data generation process, see Sect.~\ref{sec:problems}). We apply a GP with IDS for 100 generations to four different problem variants and measure how often the cases are included by IDS in the subset of relevant cases. 

Figure~\ref{fig:frequency_ids} plots the four different distributions of training cases. Figure~\ref{subfig:even_ids} is the original problem; Fig.~\ref{subfig:uneven_ids} reduces the number of samples for $x<0$ mimicking the existence of edge cases. Figs.~\ref{subfig:even_outliers_ids} and \ref{subfig:uneven_outliers_ids} are analogous with 5\% additional outliers. 
Each dot is colored and the color indicates how often this case is included in a subset with IDS over 100 generations using a down-sampling rate of $n=0.1$. We expect that IDS focuses on edge cases by including them more frequently in its subsets. This behavior is in contrast to RDS, where each case would be included with the same probability (uniformly distributed over all training cases; for the example at hand, RDS would sample each case with probability 0.1).

In Fig.~\ref{subfig:even_ids}, where all training cases are evenly distributed, IDS slightly prefers cases that are more to the left and right as well as the inflection point in the middle compared to the cases in between. The edge cases are included much more frequently (around $5$ times more often than with RDS).
In Fig.~\ref{subfig:uneven_ids}, for an uneven distribution of training cases, IDS focuses on the cases with $x<0$ in addition to the edge cases. It includes training cases from the under-represented problem region much more frequently compared to the cases lying in dense regions. This clearly highlights the advantage of IDS, as it ensures that cases covering different behaviors are included. In contrast, for RDS, cases from the under-represented problem region would have an equal chance of being sampled than the cases from the dense region ($\approx 0.1$). Thus, the important edge cases are often omitted for several generations. Consequently, RDS is not sufficiently evaluating candidate solutions on edge cases, which potentially harms the search process.

Unlike synthetic problems, real-world regression problems often contain outliers, meaning training cases that significantly deviate from the regular pattern~\cite{chandola2009anomaly}. 
In contrast to edge cases, however, they are not relevant and we want optimization methods to ignore them as their consideration impedes the search process \cite{lopez2017ransac}.
To demonstrate the behavior of IDS in the presence of outliers, we added outliers to both variants of the problem (see Figs.~\ref{subfig:even_outliers_ids} and \ref{subfig:uneven_outliers_ids}). 

In Fig.~\ref{subfig:even_outliers_ids}, for an even distribution of cases but with additional outliers, we see that IDS strongly focuses on the outliers and includes them very often in the subsets. For example, the outlier at $y=-6$ is included almost every time. In contrast, RDS would be much less affected by outliers, as the chance of being selected is uniform across all cases. Finally, Fig.~\ref{subfig:uneven_outliers_ids} illustrates for an uneven distribution with additional outliers, that the benefits of IDS are mitigated because IDS now focuses primarily on the outliers and not on the edge cases.

To sum up, the behavior of IDS to over-represent existing outliers in the subsets mitigates the positive effect of IDS to focus on the edge cases. As soon as outliers are present, IDS focuses strongly on outliers not covering the regular data pattern any more.

\section{ROIDS: Robust Outlier-Aware Informed Down-Sampling}\label{sec:method}

To mitigate the limitations of IDS in the presence of outliers we propose ROIDS, an extension of IDS, which keeps the benefits of IDS while mitigating its brittleness to outliers. 
ROIDS builds on the assumption that outliers, in contrast to edge cases, are fundamentally different from the underlying pattern~\cite{chandola2009anomaly} and hard to learn without overfitting these cases \cite{lopez2017ransac}. 
As populations of candidate solutions effectively act as ensembles, such populations are inherently less prone to overfitting these outliers \cite{kotanchek2009symbolic, lopez2018filtering, trujillo2019untapped}.
Instead, a population should have on average a high mean error on these outlier cases. 
ROIDS exploits this property by excluding the cases with the highest mean error across candidates from a possible subset at any given generation.
As a result, ROIDS is able to create diverse subsets of training cases without oversampling outliers. 

\begin{algorithm}[t]
\caption{Robust Outlier-Aware Informed Down-sampling}
\label{algo:roids}
\begin{algorithmic}[1]
\Require
Parent population $P$, training cases $T$, current generation $G$, down-sample rate $n$,
parent sampling rate $\rho$, scheduled case distance computation parameter $l$, outlier sensitivity rate $\gamma$
\Ensure
Down-sampled training set $N$

\If{$G \% l == 0$}\label{line:scheduled_case_distance}
    \State $\hat{P} \gets$ sample $\rho \cdot |P|$ candidates from $P$\label{line:parent_sampling}
    \State evaluate each candidate $\hat{p} \in \hat{P}$ on each case $t \in T$ to get \hspace*{1.3em} all case error vectors $s \in S$\label{line:evaluation}
    \State calculate mean error value of each $s\in S$ \label{line:error_calculation}
    \State $\hat{T} \gets$ $T$ after removing  $\gamma*|T|$ cases with highest mean \hspace*{1.3em} error value \label{line:remove_outliers}
    \State $D \gets$ compute distances using case error vectors of $\hat{T}$\label{line:distance_matrix}
\EndIf
\State $N \gets$ sample $n*|T|$ cases from $\hat{T}$ with farthest first traversal using $D$\label{line:create_subset}
\State \Return $N$
\end{algorithmic}
\end{algorithm}

Algorithm~\ref{algo:roids} describes the ROIDS procedure in detail.
Similar to IDS \cite{boldi2024informed}, a distance matrix between cases is calculated every $l$ generations (line $1 - 7$).
This is done by first sampling a fraction $\rho$ of the parent population $P$ (line 2). 
This sampled parent population $\hat{P}$ is then evaluated on the whole training set $T$ to construct a case error vector $s \in S$ for each case $t \in T$ (line 3), therefore, $s$ describes the non-aggregated performance of the sampled parent population on a specific case.
For example, if $T$ contains $100$ training cases and $\hat{P}$ a sample of $5$ candidate solutions, $S$ contains $100$ case error vectors $s$ of size $5$. 

Contrary to the original IDS algorithm, ROIDS adds an additional step at this point: Prior to the construction of the down-sample, ROIDS identifies possible outliers by collapsing each case error vector to its mean error value and masking those cases from inclusion in the down-sample with the highest mean error (line $4 - 5$). Specifically, $\gamma * |T|$ cases are excluded from $T$ resulting in $\hat{T}$, where $\gamma$ is the adjustable outlier sensitivity rate in $[0,1)$. This additional step of ROIDS comes at almost no additional cost in comparison to the original IDS algorithm. 

Afterwards, the ROIDS algorithm continues as the original IDS algorithm would: A distance matrix $D$ is constructed using the pairwise distances between all cases $\hat{t} \in \hat{T}$ with regards to their case error vectors (line 6).
This distance matrix $D$ is then used each generation to construct a subset $N$ of size $n*|T|$ using the farthest first traversal algorithm \cite{hochbaum1985best} to include the most informative cases from $\hat{T}$ (line 8) .

Finally, the subset $N$ is used to evaluate the quality of the candidate solutions in a population. By including diverse and highly informative cases, the candidates are sufficiently evaluated on different problem regions. Further, by excluding potential outliers from the subset, ROIDS guides the search toward a meaningful direction instead of overfitting those outliers.

\section{Experimental Setup}\label{sec:experimental_setup}

This section describes the used benchmark problems as well as the GP setup and parameter settings.

\subsection{Problems}\label{sec:problems}
We first run experiments on synthetic problems that can be manipulated to contain outliers. This controlled setup allows us to test our assumption that IDS performs poorly in the presence of outliers, and to show that ROIDS is more robust in such situations. 

 First, we study the behavior of the down-sampling methods for four variants of the $2$D \texttt{nguyen-6} problem~\cite{uy2011semantically}, which is defined as 
\begin{equation}\label{nguyen6}
y = sin(x) + sin(x + x^2).
\end{equation}

For the first variant with an even distribution, 100 data points were placed evenly across $[-1,1]$. For the second variant with an uneven distribution, 10\% of the data points were placed in $[-1,0]$ and 90\% in $[0,1]$. Additionally, we generated two datasets (with even and uneven distribution) containing outliers by corrupting 5\% of the data points with noise from a Gaussian distribution with a mean of zero and a standard deviation four times that of the original $y$-values.
We use the same test set for all variants of the \texttt{nguyen-6} problem, which contains 100 evenly placed data points between $[-1,1]$ and is outlier free. 

In addition to that, we generated the \texttt{friedman1}, \texttt{friedman2}, and \texttt{friedman3} datasets as described in~\cite{friedman1991multivariate, breiman1996bagging} with 100 training points and 100 test points. For each of the datasets we also generated a variant containing outliers by manipulating 5\% of the data points as described before. Again, the test set is outlier free.

Additionally, we test the performance of ROIDS on a selection of $6$ real-world problems that are commonly used in literature~\cite{geiger.2023, LaCava.2019, Virgolin.2021}. For these problems, we randomly sampled $15\%$ of the data points as a test set and the rest for training. The dimension of each problem, as well as the number of training and test cases is listed in Table~\ref{tab:problems}.

\begin{table}
    \caption{Benchmark problems with their dimension, number of training and test cases and the problem type.}
    \centering
    \begin{tabular}{lrrrr}
         \toprule
         Problem  & Dimension & \#Train & \#Test & Type\\
         \midrule
         \texttt{nguyen-6} \cite{uy2011semantically}  & 1 & 100& 100 & synthetic\\
         \texttt{friedman1} \cite{friedman1991multivariate, breiman1996bagging} & 5 & 100& 100 & synthetic\\
         \texttt{friedman2} \cite{friedman1991multivariate, breiman1996bagging}& 4 & 100& 100 & synthetic \\
         \texttt{friedman3} \cite{friedman1991multivariate, breiman1996bagging}  & 4 & 100& 100 & synthetic\\
         \texttt{airfoil} \cite{Brooks.1989} & 5 & 1277 & 226 & real-world\\
         \texttt{concrete} \cite{Yeh.1998}  & 8 & 875 & 155&real-world\\
         \texttt{enh} \cite{Tsanas.2012} & 8 & 652 & 116 & real-world \\
         \texttt{housing} \cite{Harrison.1978} & 13 & 430 & 76&real-world\\
         \texttt{redwine} \cite{cortez2009modeling} & 11 & 1359 & 240 &real-world\\
         \texttt{yacht}  \cite{Gerritsma.1981} & 6 & 261& 47&real-world\\
         \bottomrule
    \end{tabular}
    \label{tab:problems}
\end{table}

The number of training cases varies between $261$ for the \texttt{yacht} problem to $1,359$ for the \texttt{redwine} problem. The number of features goes up to $13$ for the \texttt{housing} problem.

\subsection{Parameter Setting}

For our experiments, we used the DEAP framework~\cite{Fortin.2012} as foundation of our implementation. The GP approach uses standard parameter settings described in Table~\ref{tab:parameter_setting}.

\begin{table}
  \centering
\caption{Parameter settings of our GP approach.}
\label{tab:parameter_setting}
\begin{tabular}{l|r}
 \toprule 
\textbf{Parameter} & \textbf{Value} \\
\midrule
Population size & $500$ \\
Generations & 500 (100 for \texttt{nguyen-6}) \\
Primitive set & $\{\textrm{\textbf{x}}, \textrm{ERC}, +, -, *, \textrm{AQ}~\cite{Ni.2013}, \textrm{sin}, \textrm{cos}, \textrm{neg}\}$ \\
ERC values & $\{-1,0,1\}$ \\
Selection method & Tournament ($n=7$)\\
Initialization method & Ramped half-and-half \\
Maximum tree depth & $17$ \\
Crossover probability & $80\%$ \\
Mutation probability & $5\%$ \\
Down-sampling rate $n$ & $0.1$ \\
Runs & 100 \\
\bottomrule
\end{tabular}
\end{table}

We initialized the population using ramped half-and-half and set the population size to $500$. The crossover and mutation probabilities are $80\%$ and $5\%$, respectively. We run our experiments for $500$ generations for the Friedman and real-world problems. For the \texttt{nguyen-6} problem, we set the generation limit to $100$ as it is a much simpler problem. We use  tournament selection, because previous work found that for symbolic regression problems tournament selection performs similarly to lexicase selection when combined with down-sampling methods, but runs faster~\cite{geigertournament}. We set the tournament size to $7$. The quality of a candidate solution is measured as its mean squared error (MSE) on the subset of training cases.

For all down-sampling methods, we define a down-sampling rate of $n=0.1$~\cite{geiger.2023}, meaning that only 10\% of the training cases are used in each generation to evaluate the quality of the candidate solutions. For IDS and ROIDS, we set the parent sampling rate to $\rho=0.01$ and the scheduled case distance computation parameter to $l=10$ as suggested by~\citet{boldi2024informed}. 
For ROIDS, the outlier sensitivity rate is set to $\gamma=0.05$ (an ablation study with results for different $\gamma$ are shown in Figs.~\ref{fig:friedman_gammas} and \ref{fig:realworld_gammas} of the appendix).

Additionally, to further motivate the design of ROIDS, we included a baseline in our real-world experiments that filters outliers from the training set before starting an evolutionary run with IDS. To remove the outliers, we used the Local Outlier Factor (LOF)~\cite{breunig2000lof}, which is a widely used outlier detection method~\cite{cheng2019outlier}. We implemented LOF with its default parameters in scikit-learn~\cite{scikit-learn}. 

For each run, we randomly take 15\% of the training cases as a validation set. The validation set is used to choose the final solution from all candidates during a run (meaning the candidate with the lowest MSE on the validation set). This candidate is then evaluated on the unseen test set. 

All experiments are repeated $100$ times. For the analysis of the parameter $\gamma$ (see supplementary material), we performed $30$ repetitions.

\section{Results}\label{sec:results}

Section~\ref{sec:results_synthetic} presents an analysis of the behavior and performance of ROIDS on synthetic regression problems. In Sect.~\ref{sec:results_realworld}, we study whether our observations transfer to real-world regression datasets. 

\subsection{Performance of ROIDS on Synthetic Regression Problems}\label{sec:results_synthetic}

\begin{figure}
    \centering
    \begin{subfigure}[b]{0.235\textwidth}
        \centering
        \includegraphics[width=\linewidth]{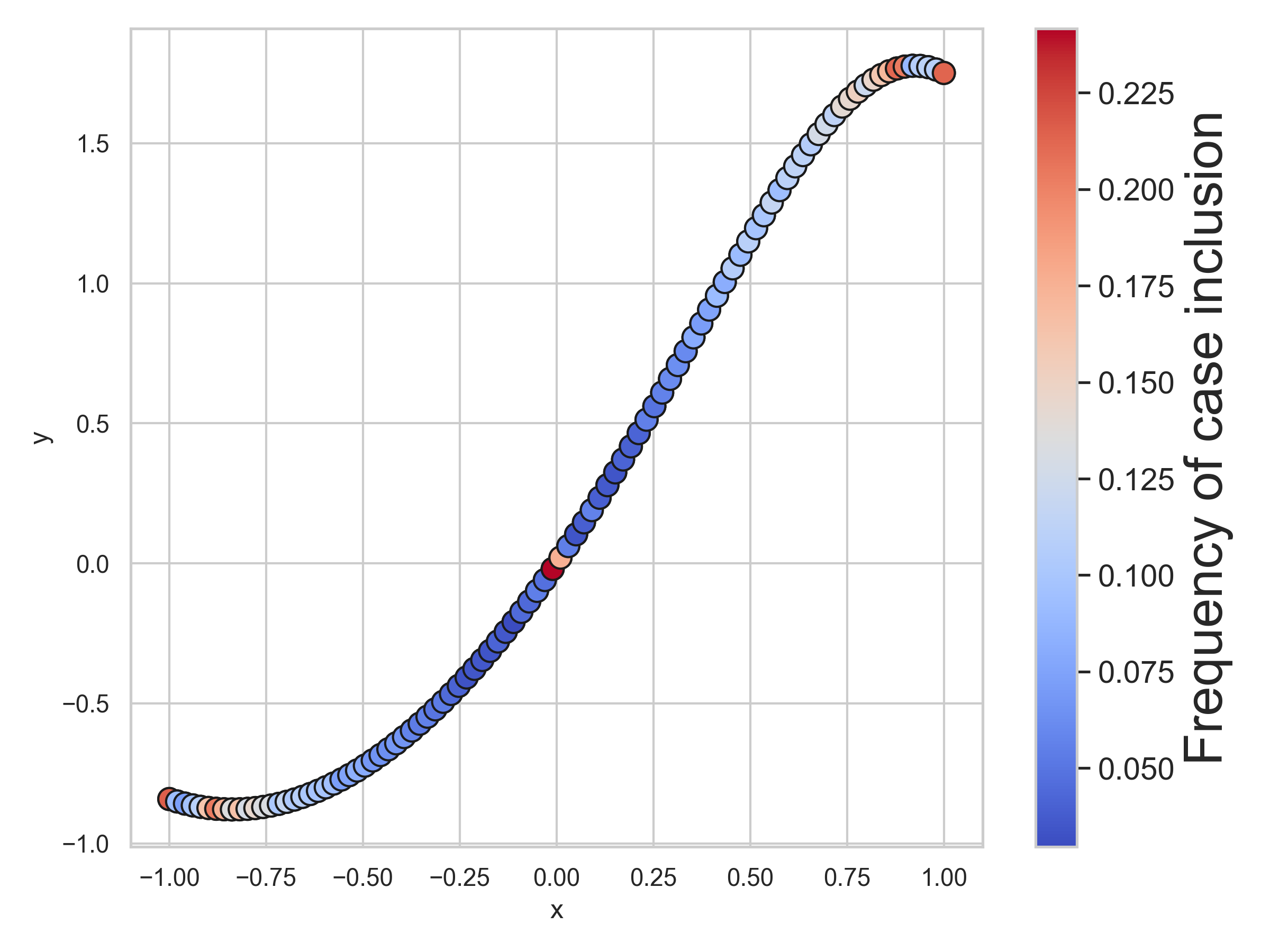}
\caption{Even distribution of \\training cases.}
        \label{subfig:even_roids}
    \end{subfigure}
    \hfill
    \begin{subfigure}[b]{0.235\textwidth}
        \centering
        \includegraphics[width=\linewidth]{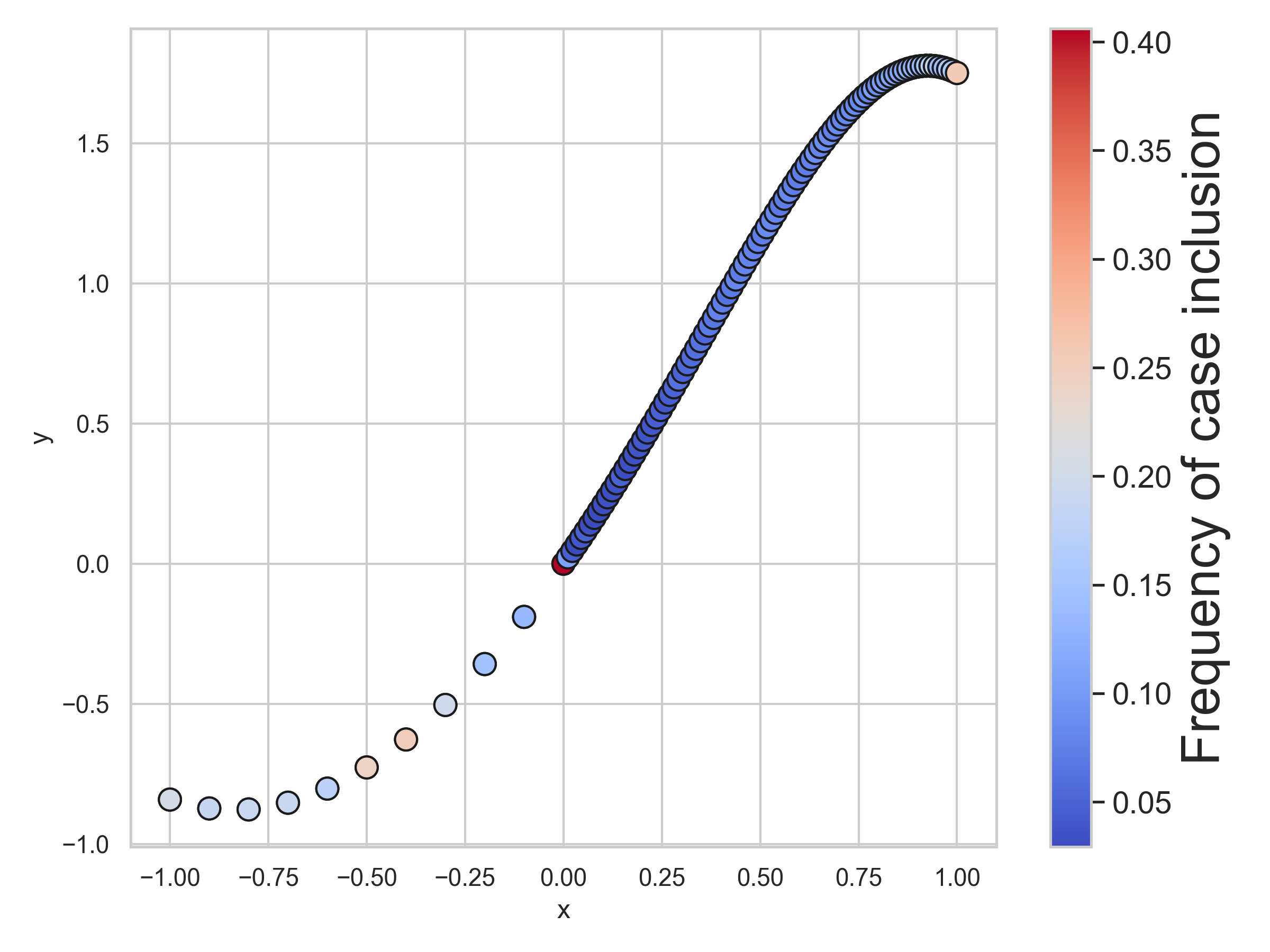}
        \caption{Uneven distribution of \\training cases.}
        \label{subfig:uneven_roids}
    \end{subfigure}
    \begin{subfigure}[b]{0.235\textwidth}
        \centering
        \includegraphics[width=\linewidth]{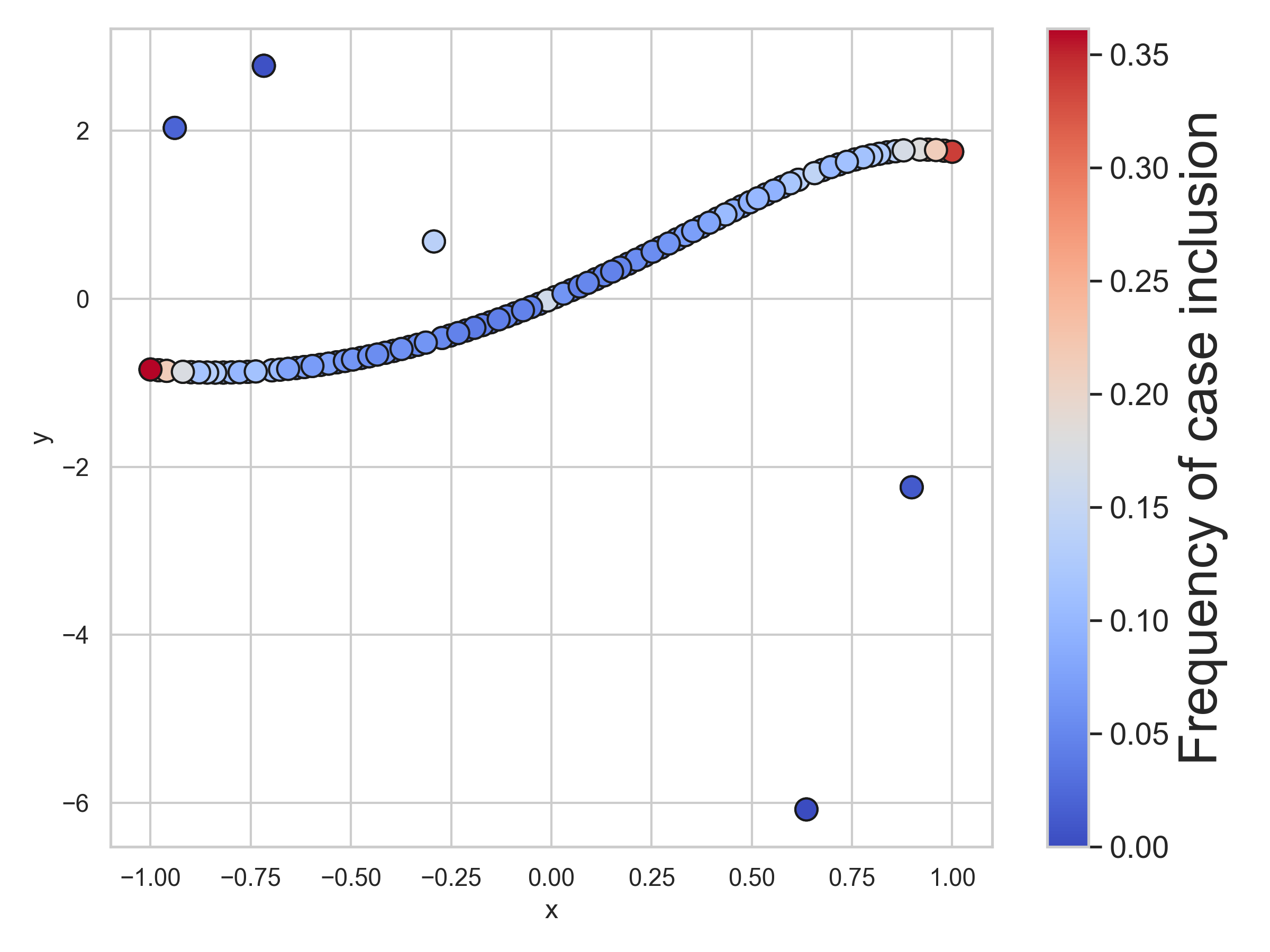}
        \caption{Even distribution of \\training cases and 5\% outliers.}
        \label{subfig:even_outliers_roids}
    \end{subfigure}
    \hfill
    \begin{subfigure}[b]{0.235\textwidth}
        \centering
        \includegraphics[width=\linewidth]{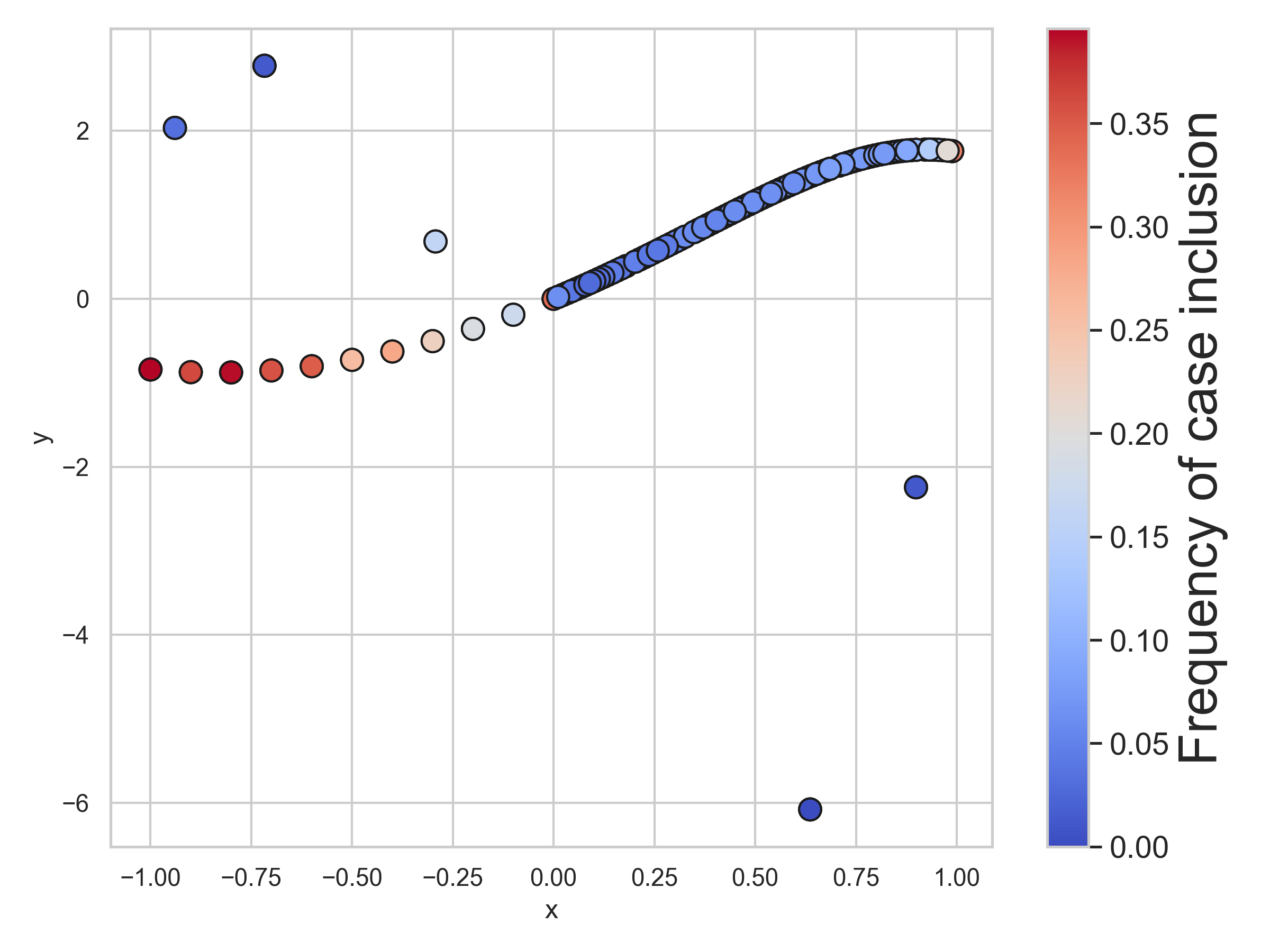}
        \caption{Uneven distribution of \\training cases and 5\% outliers.}
        \label{subfig:uneven_outliers_roids} 
    \end{subfigure}
    \caption{Color-coded frequencies of including a training case in the subsets when using ROIDS. Results are for different variants of the \texttt{nguyen-6} problem. The top row contains two outlier-free variants of the problem, whereas the bottom row includes 5\% outliers.}
    \label{fig:roids_nguyen}
\end{figure}

Section~\ref{sec:analysis} exemplarily illustrated for the \texttt{nguyen-6} problem that IDS frequently includes outliers in its subsets and, consequently, is not sufficiently covering the true underlying pattern anymore. 
Based on this observation, Sect.~\ref{sec:method} proposes ROIDS, an outlier-aware variant of IDS that excludes potential outliers from its subsets. 
In analogy to Sect.~\ref{sec:analysis} which illustrates the behavior of IDS, Fig.~\ref{fig:roids_nguyen} visualizes the frequency of case inclusion using ROIDS for the \texttt{nguyen-6} problem with different distributions of training cases and in the presence of outliers. 

For an even distribution of training cases (Figs.~\ref{subfig:even_roids} and \ref{subfig:even_ids}), ROIDS behaves similar to IDS including the cases on the edges and the inflection point more frequently compared to RDS (for RDS each case has a probability of approx.~$10\%$ to be included for a down-sampling rate of $n=0.1$). For an uneven distribution of training cases (Figs.~\ref{subfig:uneven_roids} and \ref{subfig:uneven_ids}), the situation is similar: ROIDS includes cases from the under-represented problem region ($x < 0$) as well as edge cases more frequently than RDS. This indicates that ROIDS keeps the advantages of IDS, as it includes highly informative cases more frequently and the subsets consist of cases that cover diverse behaviors. 

The advantage of ROIDS over IDS for problems with outliers becomes evident in Figs.~\ref{subfig:even_outliers_roids} and \ref{subfig:uneven_outliers_roids}: in the presence of outliers, ROIDS includes the outliers rarely in its subsets, while IDS includes them almost every generation (compare Figs.~\ref{subfig:even_outliers_ids} and ~\ref{subfig:uneven_outliers_ids}).
By excluding the outliers, ROIDS is able to focus on the true underlying data pattern and still creates meaningful subsets by including edge cases and cases from under-represented problem regions more frequently. In contrast, IDS fails to sufficiently represent the edge cases in its subsets in the presence of outliers.

\begin{figure}[t]
    \centering
    \includegraphics[width=1.0\linewidth]{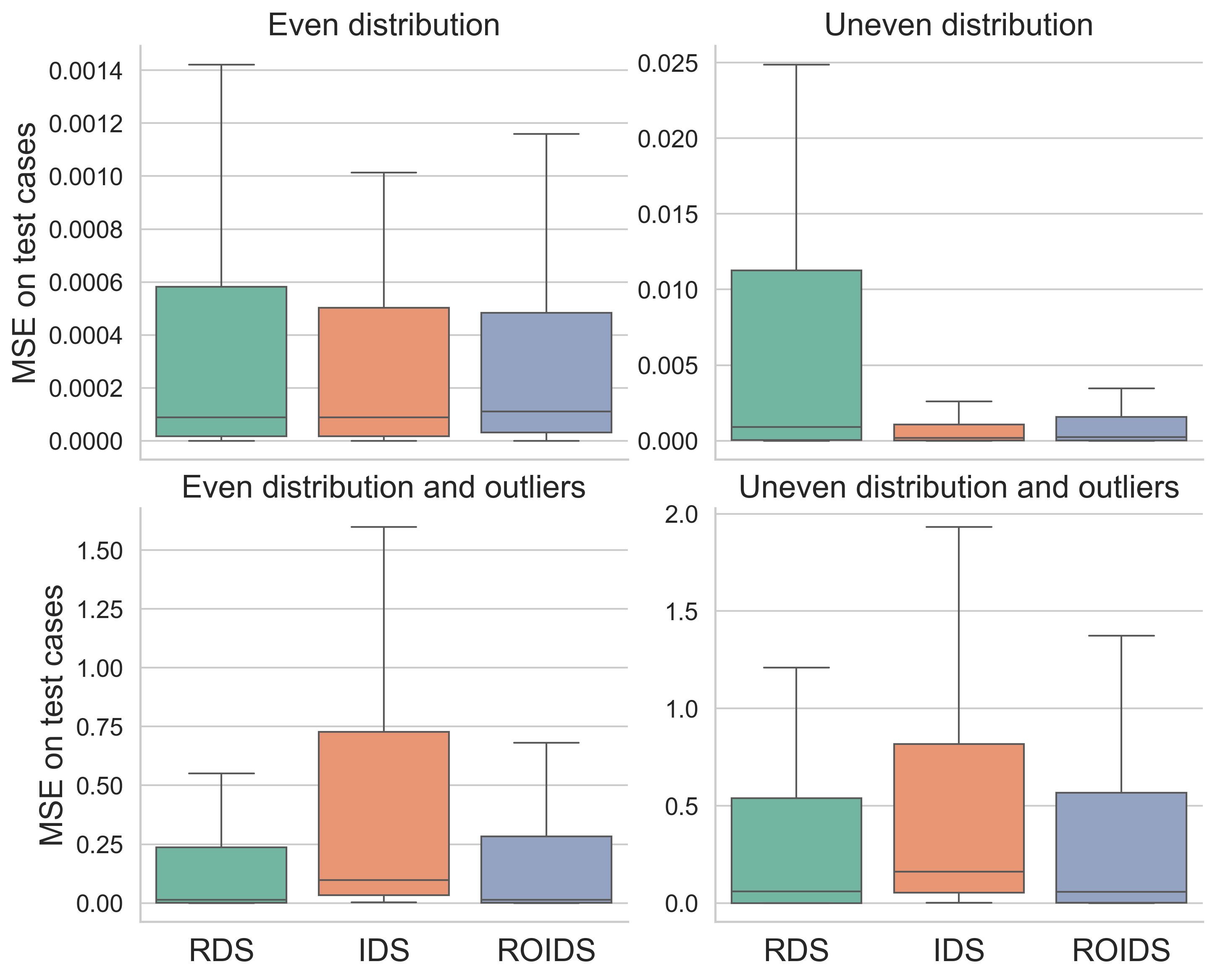}
    \caption{Performance of different down-sampling methods for different variants of the \texttt{nguyen-6} problem. Note that the y-axes of the plots are scaled differently. For better readability, outliers are not plotted.}
    \label{fig:results_nguyen}
\end{figure}

Next, we analyze how the creation of subsets affects the performance for the different variants of the \texttt{nguyen-6} problem. For the best solutions found for the validation cases, Figure~\ref{fig:results_nguyen} shows the distribution of the MSE on the test cases over 100 runs for RDS, IDS, and ROIDS. As expected, for an even distribution without outliers (top left), all down-sampling strategies perform equally well. Intuitively, if the distribution is even, RDS is also able to sufficiently cover the problem in its subsets as all training cases are equally likely to be included. 
For an uneven distribution without outliers (top right), both IDS and ROIDS achieve a lower MSE and have less variance in their results compared to RDS. The subsets created by IDS and ROIDS still cover all problem regions equally well even if the original dataset has an uneven distribution. With RDS, the under-represented problem part is also under-represented in its subsets, which harms performance. 

The situation is different when adding outliers to the  \texttt{nguyen-6} problem (bottom row of Fig.~\ref{fig:results_nguyen}). Then, IDS performs worse than RDS, because it heavily focuses on the outlier cases. In contrast, ROIDS still performs well and similar to RDS. This highlights that ROIDS successfully mitigates the drawbacks of IDS in the presence of outliers but is still able to exploit its benefits if the distribution of training cases is uneven. This makes ROIDS a good choice across a wide range of problems.

\begin{figure}
    \centering
    \includegraphics[width=1.0\linewidth]{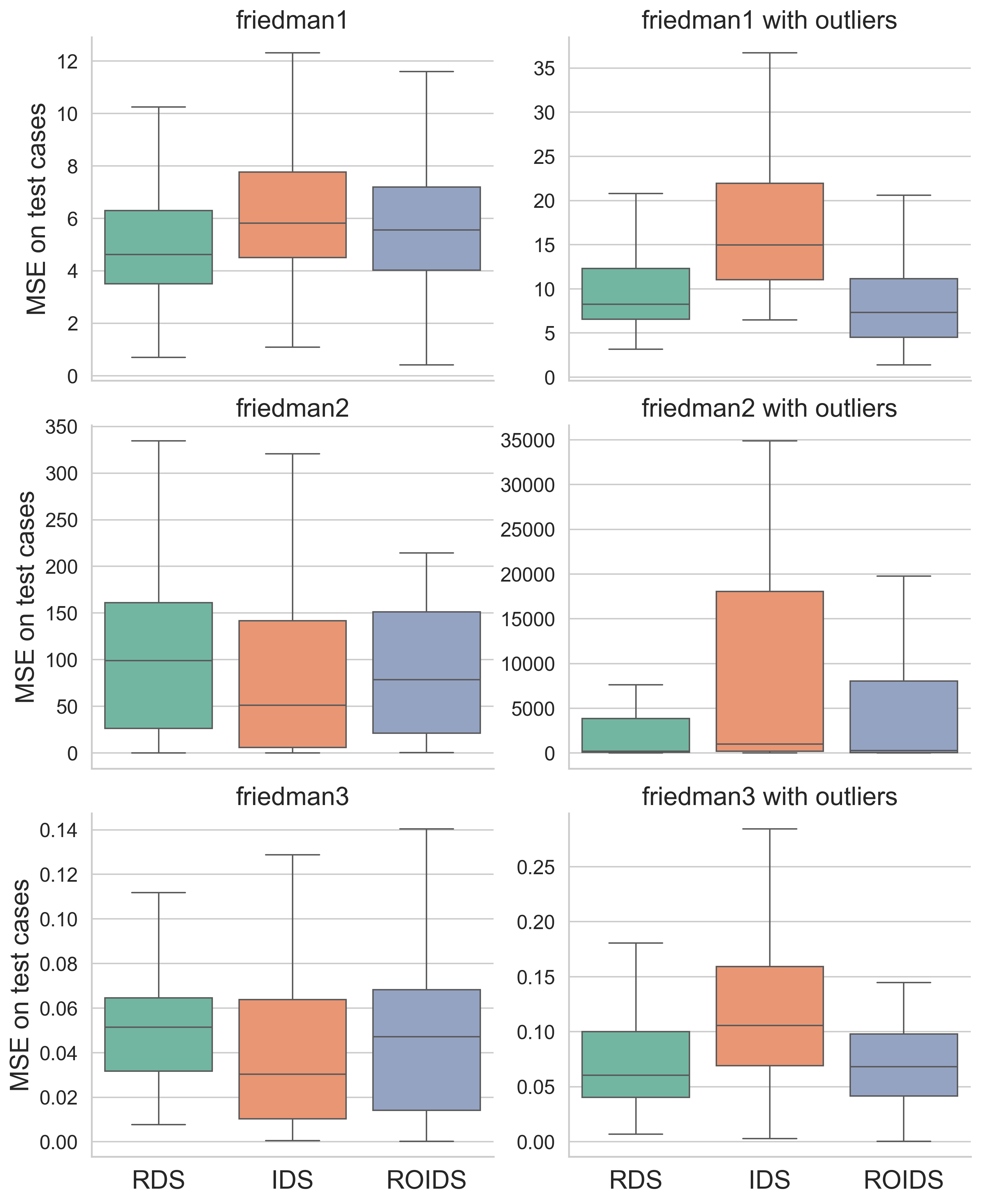}
    \caption{Performance of different down-sampling methods for Friedman problems with and without outliers. Note that the y-axes of the plots are scaled differently. For better readability, outliers are not plotted.}
    \label{fig:results_friedman}
\end{figure}

Next, we study the performance of ROIDS in comparison to RDS and IDS for three additional more complex synthetic problems, namely the \texttt{friedman1}, \texttt{friedman2}, and \texttt{friedman3} problems with and without outliers. 
Fig.~\ref{fig:results_friedman} shows the MSE on the test cases of the best found solutions for 100 runs.
Although IDS performs better than RDS on two out of three Friedman problems without outliers, it performs always worse than RDS in the presence of outliers. In contrast, ROIDS performs reasonably well across all variants of the Friedman problems. Especially in the presence of outliers, ROIDS clearly outperforms IDS on all problems.  

Table~\ref{tab:results_friedman} summarizes the median MSE on the test cases for RDS, IDS, and ROIDS for all synthetic problems. Further, we tested for statistically significant differences between ROIDS and IDS using the Mann-Whitney U test~\cite{mann1947test}, with the significance threshold set to $\alpha = 0.05$.
ROIDS achieves the lowest MSE on 4 out of 10 problems and the second best on the remaining problems. In contrast to RDS and IDS, ROIDS is never the worst performing method. This is also reflected in the mean ranking of each method. IDS achieves the worst mean ranking with 2.2, RDS the second best with 1.8, and ROIDS the best with 1.6. This confirms not only that ROIDS performs well across different problems but, moreover, that ROIDS always performs significantly better than IDS if outliers are present and never significantly worse otherwise.

For the interested reader, we visualized the structure of the Friedman problems with outliers using UMAPs for dimensionality reduction in Appendix \ref{appendix:friedman} (see Figs.~\ref{fig:umap_friedman1}-~\ref{fig:umap_friedman3}) to show the frequency of case inclusion for the different down-sampling methods. We find that IDS includes the outlier cases very frequently while ROIDS mostly excludes them from the subset creation. 

Finally, we study how the performance of ROIDS depends on the setting of the outlier sensitivity rate $\gamma$. In the supplementary material (Appendix~\ref{appendix:friedman}), Fig.~\ref{fig:friedman_gammas} shows how the performance changes for $\gamma \in \{0.025, 0.05, 0.1\}$. We find that for Friedman problems without outliers a smaller $\gamma$ performs slightly better. In the presence of outliers, a $\gamma$ smaller than the amount of added outliers (here $5\%$) performs slightly worse. 

\begin{table}
    \centering
        \caption{Performance of different down-sampling methods for synthetic problems with and without outliers. The median MSE on the test cases over 100 runs is shown. Best results are highlighted in bold. Significant differences between ROIDS and IDS are indicated by an asterisk.}
\begin{tabular}{lrr|r}
\toprule
Problem &    RDS &     IDS &  ROIDS\\
\midrule
\texttt{nguyen-6} even & \textbf{0.000} & \textbf{0.000} & \textbf{0.000} \\
\texttt{nguyen-6} even + outliers & \textbf{0.013} & 0.097 & $^*$0.014 \\
\texttt{nguyen-6} uneven & 0.001 & \textbf{0.000} & \textbf{0.000} \\
\texttt{nguyen-6} uneven + outliers & 0.059 & 0.160 & $^*$\textbf{0.058} \\
\texttt{friedman1} & \textbf{4.625} & 5.824 & 5.558 \\
\texttt{friedman1} + outliers & 8.273 & 14.955 & *\textbf{7.318} \\
\texttt{friedman2} & 98.886 & \textbf{51.320} & 78.664 \\
\texttt{friedman2} + outliers & \textbf{209.587} & 994.823 & *261.759  \\
\texttt{friedman3} & 0.051 & \textbf{0.030} & 0.047 \\
\texttt{friedman3} + outliers & \textbf{0.060} & 0.106 & *0.068 \\
\midrule
Mean Ranking & 1.8 & 2.2 & \textbf{1.6} \\
\bottomrule
\end{tabular}
    \label{tab:results_friedman}
\end{table}

To sum up, in the presence of outliers, ROIDS is able to focus on the true underlying pattern and consequently, guides the search process more successfully than IDS. We found that ROIDS always performs significantly better than IDS if outliers are present. 
Moreover, the mean ranking of ROIDS across all synthetic problems is better compared to RDS and IDS, meaning ROIDS is robust and performs well across a wider range of problems.

\subsection{Performance of ROIDS on Real-World Regression Problems}\label{sec:results_realworld}

This subsection  analyzes the performance of ROIDS on real-world regression problems to test if the performance benefits observed in Sect.~\ref{sec:results_synthetic} also transfer to real-world settings. To motivate our design choice of ROIDS, we include an additional baseline variant called LOF+IDS, which excludes outliers using LOF before performing an evolutionary run with IDS.

\begin{figure}
    \centering
    \includegraphics[width=1.0\linewidth]{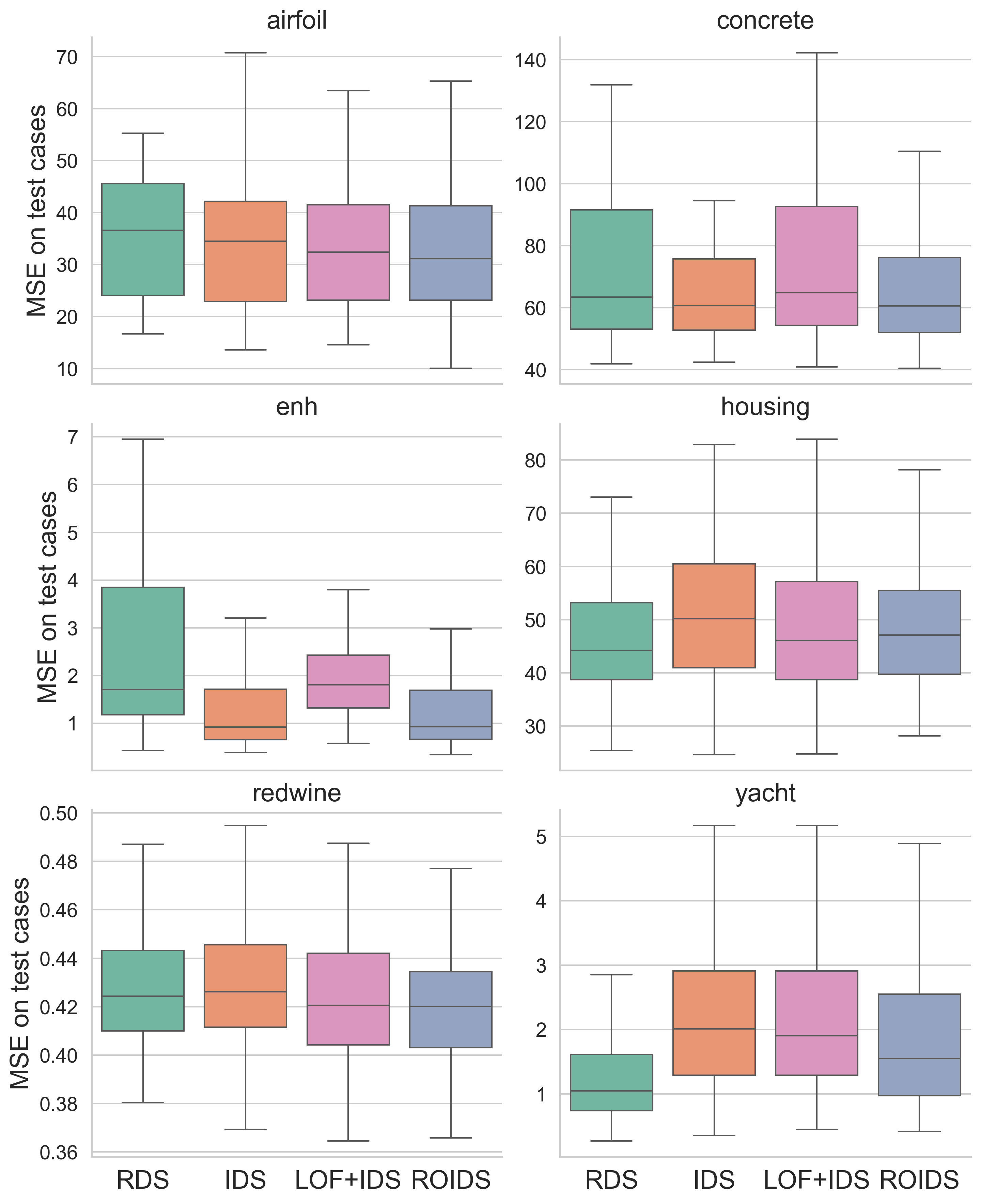}
    \caption{Performance of different down-sampling methods for real-world problems.  For better readability, outliers are not plotted.}
    \label{fig:results_realworld}
\end{figure}

Figure~\ref{fig:results_realworld} compares the performance of RDS, IDS, LOF+IDS, and ROIDS for six real-world regression problems over 100 runs. Further, Table~\ref{tab:results_realworld} summarizes the median MSE on the test cases of the best found solutions for all down-sampling methods. 
We find that ROIDS mostly achieves a lower median MSE compared to LOF+IDS. This motivates our design choice for dynamically calculating the outliers every $l$ generations based on the average error across the population on each case. In contrast to ROIDS, LOF removes outliers by comparing the local density of each training case to the local density of its $k$-nearest neighbors~\cite{breunig2000lof}. However, the interpretation of local density is difficult as it depends on the given data distribution~\cite{kriegel2011interpreting}, which might also lead to the exclusion of important edge cases. Therefore, we find that ROIDS is better in distinguishing between edge cases and outliers in comparison to LOF+IDS. 

For the \texttt{housing}, \texttt{redwine}, and \texttt{yacht} problem, IDS achieves a worse median MSE than RDS. Here, the median MSE of ROIDS is always better than that of IDS and ROIDS even significantly outperforms IDS on the \texttt{redwine} and \texttt{yacht} problem. 

Only on the \texttt{enh} problem, IDS performs best, and notably better than RDS. RDS performs best on two problems, and ROIDS is the best performing method on three problems. Again, ROIDS is never the worst performing method. This is also reflected in its mean ranking of 1.7, which is much better compared to RDS with 2.5, IDS with 3, and LOF+IDS with 2.8. 

As for the synthetic problems, we analyze the influence of the outlier sensitivity rate $\gamma$ on ROIDS performance in the supplementary material (App.~\ref{appendix:real_world}). Figure~\ref{fig:realworld_gammas} shows the performance of ROIDS for $\gamma \in \{0.025, 0.05, 0.1\}$ on all real-world problems. We can see that for real-world problems the performance of ROIDS is robust over  a reasonable range of $\gamma$ and $\gamma=0.05$ performs well across all considered problems. 

\begin{table}
    \centering
    \caption{Performance of different down-sampling methods for real-world problems. The median MSE on the test cases over 100 runs is shown. Best results are highlighted in bold. Significant differences between ROIDS and IDS are indicated by an asterisk.}
\begin{tabular}{lrrr|r}
\toprule
Problem &   RDS &  IDS & LOF+IDS & ROIDS \\
\midrule
\texttt{airfoil} & 36.599 & 34.462 & 32.412 & \textbf{31.132} \\
\texttt{concrete} & 63.467 & 60.646 & 64.826 & \textbf{60.531} \\
\texttt{enh} & 1.709 & \textbf{0.921} & 1.804 & 0.926 \\
\texttt{housing} & \textbf{44.202} & 50.183 & 46.105 & 47.080 \\
\texttt{redwine} & 0.424 & 0.426 & 0.421 & $^*$\textbf{0.420} \\
\texttt{yacht} & \textbf{1.047} & 2.010 & 1.904 & $^*$1.550 \\
\midrule
Mean Ranking & 2.5 & 3 & 2.8 & \textbf{1.7}\\
\bottomrule
\end{tabular}
    \label{tab:results_realworld}
\end{table}

\begin{figure}
    \centering
    \begin{subfigure}[b]{0.208\textwidth}
        \centering
        \includegraphics[width=\linewidth]{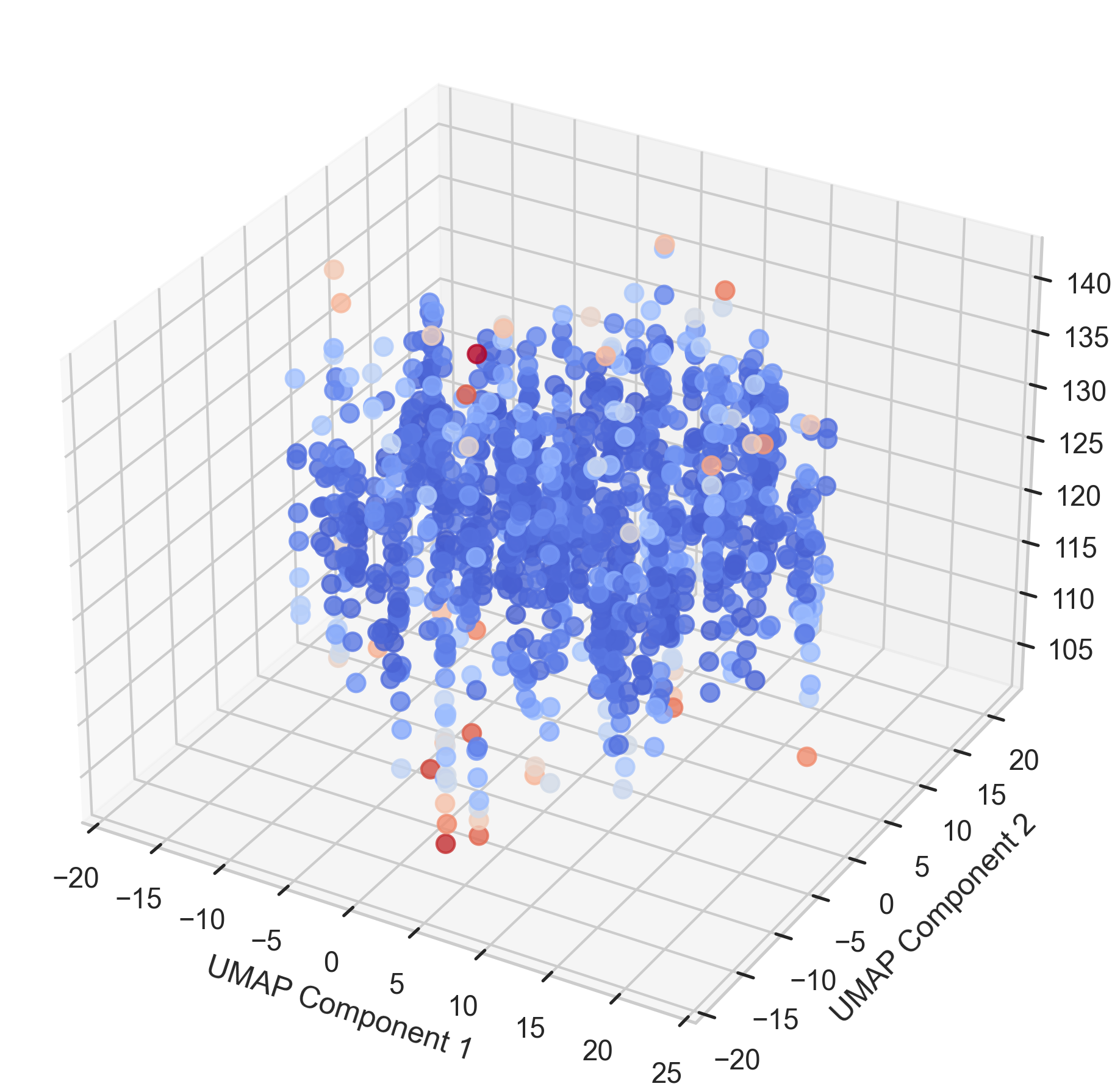}
\caption{IDS}
        \label{fig:umap_airfoil_ids}
    \end{subfigure}%
    \hfill
    \begin{subfigure}[b]{0.268\textwidth}
        \centering
        \includegraphics[width=\linewidth]{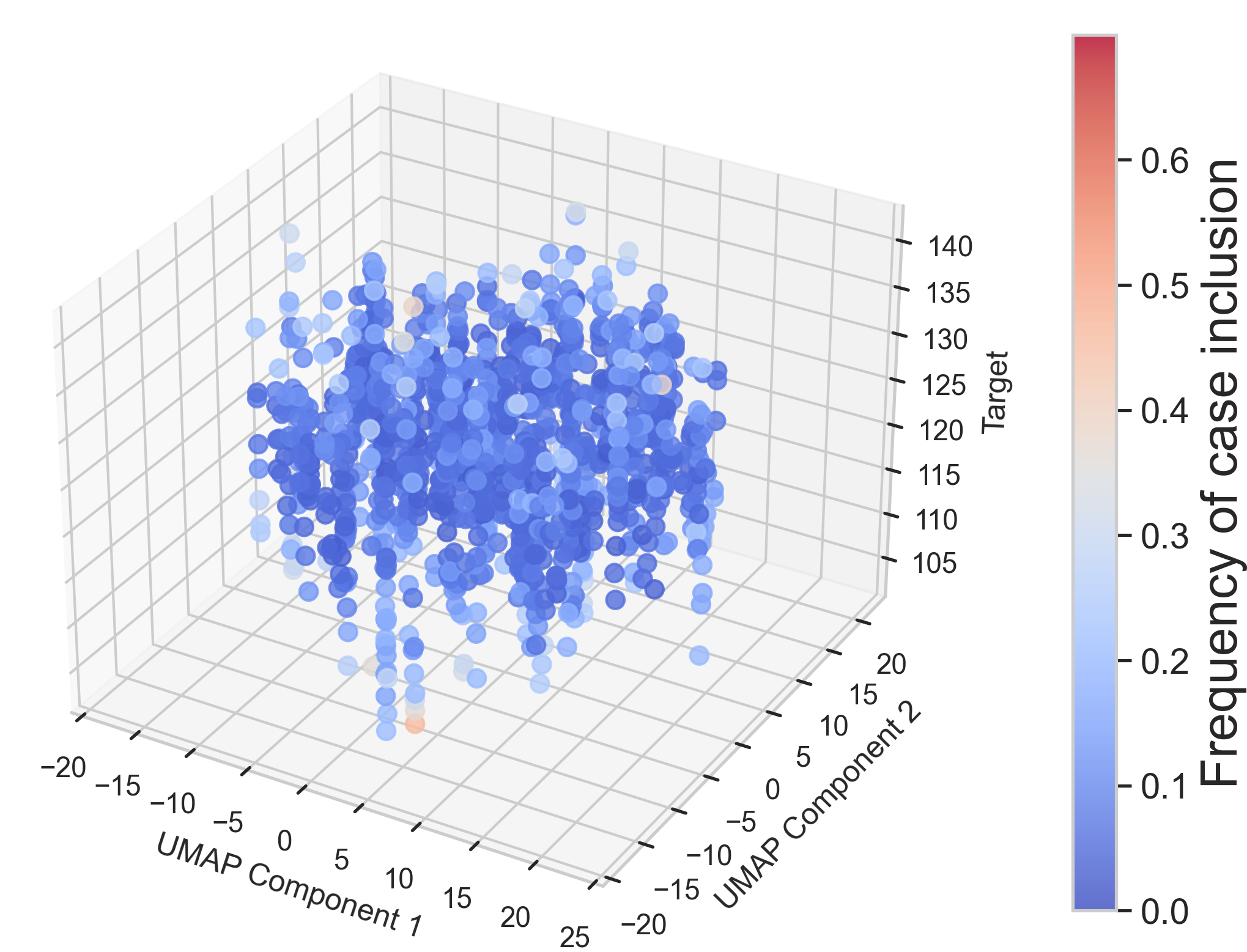}
        \caption{ROIDS}
        \label{fig:umap_airfoil_roids}
    \end{subfigure}
    \caption{UMAP Visualization of the \texttt{airfoil} problem. The color scale incidates how often each case is selected using IDS and ROIDS, respectively. The training cases are evenly distributed, which explains that all down-sampling methods perform equally well on this problem.}
    \label{fig:umap_airfoil}
\end{figure}

Additionally, we visualize the frequency of case inclusion with IDS and ROIDS for the real-world problems using UMAP as dimensionality reduction technique. For the \texttt{airfoil} problem (Fig.~\ref{fig:umap_airfoil}), we see that all training cases lie in one dense region and that there are no visual outliers present. This probably explains why the performance differences between the down-sampling methods are minimal for \texttt{airfoil}. Although IDS focuses on edge cases (see Fig.~\ref{fig:umap_airfoil_ids}), RDS probably also covers all problem regions well due to the even distribution of training cases. ROIDS includes the edge cases less frequently compared to IDS (see Fig.~\ref{fig:umap_airfoil_roids}). 

\begin{figure}
    \centering
    \begin{subfigure}[b]{0.208\textwidth}
        \centering
        \includegraphics[width=\linewidth]{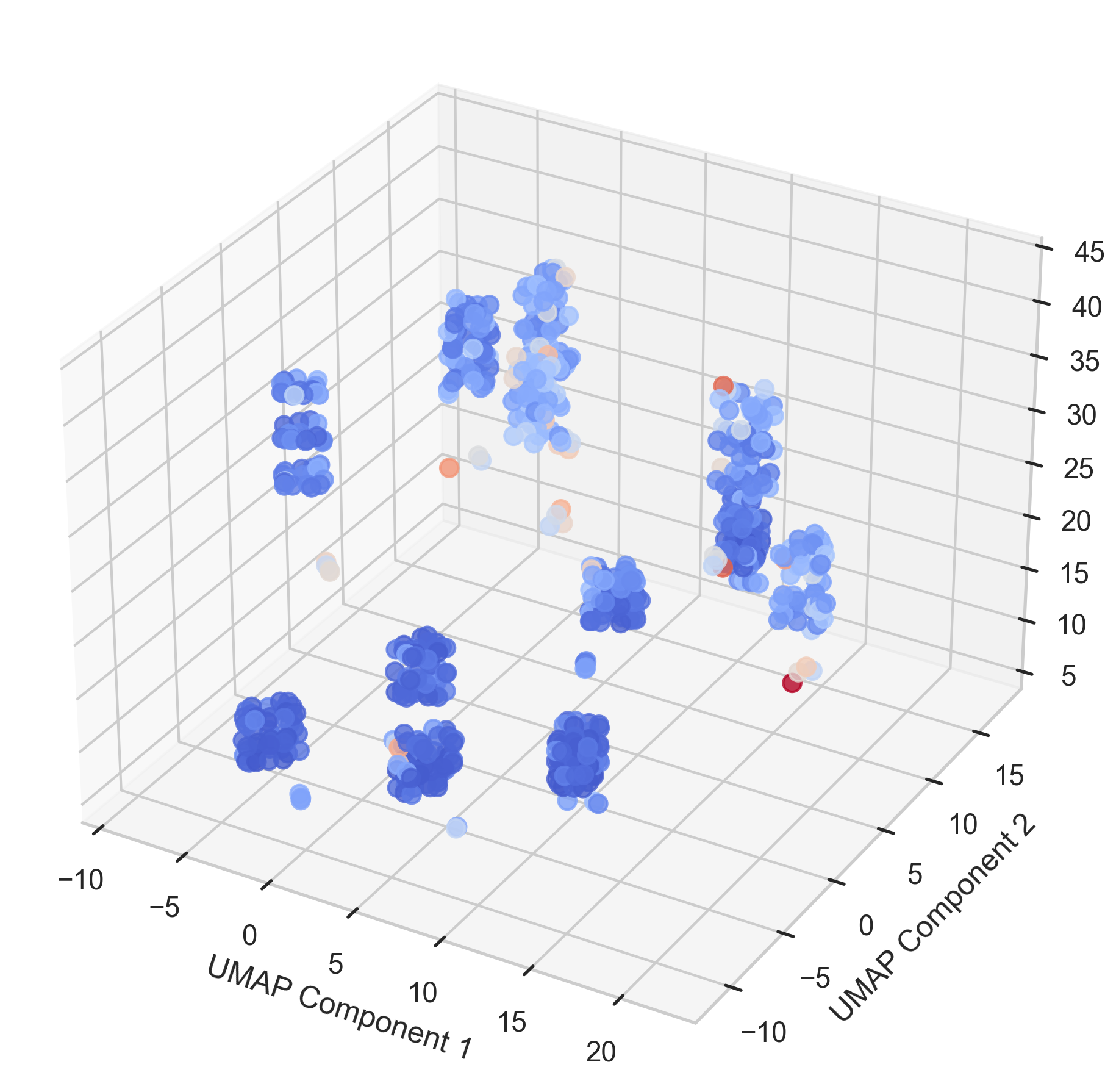}
\caption{IDS}
        \label{fig:umap_enh_ids}
    \end{subfigure}%
    \hfill
    \begin{subfigure}[b]{0.268\textwidth}
        \centering
        \includegraphics[width=\linewidth]{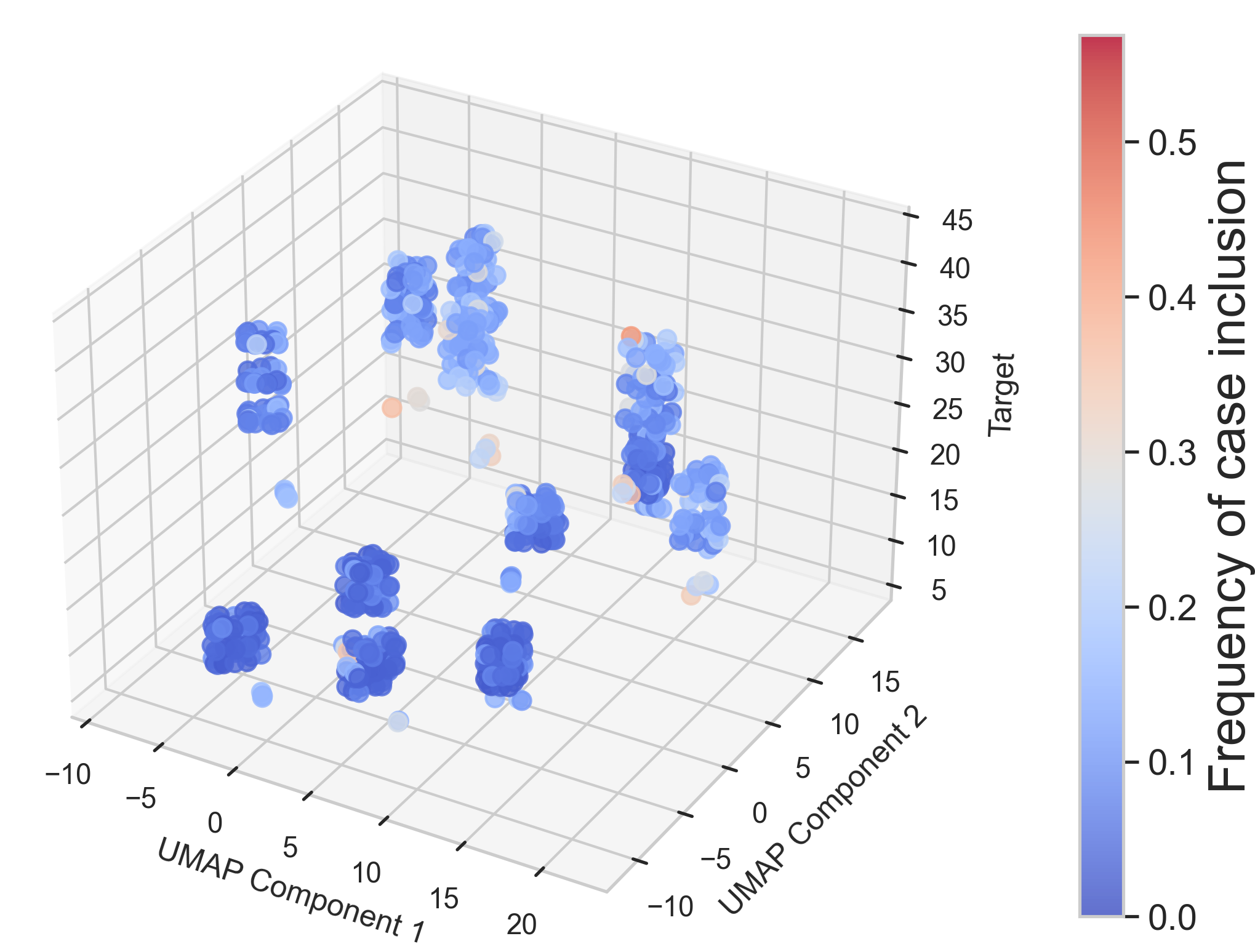}
        \caption{ROIDS}
        \label{fig:umap_enh_roids}
    \end{subfigure}
    \caption{UMAP Visualization of the \texttt{enh} problem. The color scale indicates how often each case is selected under IDS and ROIDS, respectively. The training cases are unevenly distributed across different clusters, which explains why IDS and ROIDS outperform RDS on this problem.}
    \label{fig:umap_enh}
\end{figure}

For \texttt{enh} (Fig.~\ref{fig:umap_enh}), we see that there are different clusters. Here, IDS and ROIDS frequently include edge cases from the clusters in their subsets, which probably leads to a better coverage of the problem in their subsets compared to the subsets created with RDS. This is in line with the finding, that IDS and ROIDS outperform RDS on the \texttt{enh} problem.

\begin{figure}[h]
    \centering
    \begin{subfigure}[b]{0.208\textwidth}
        \centering
        \includegraphics[width=\linewidth]{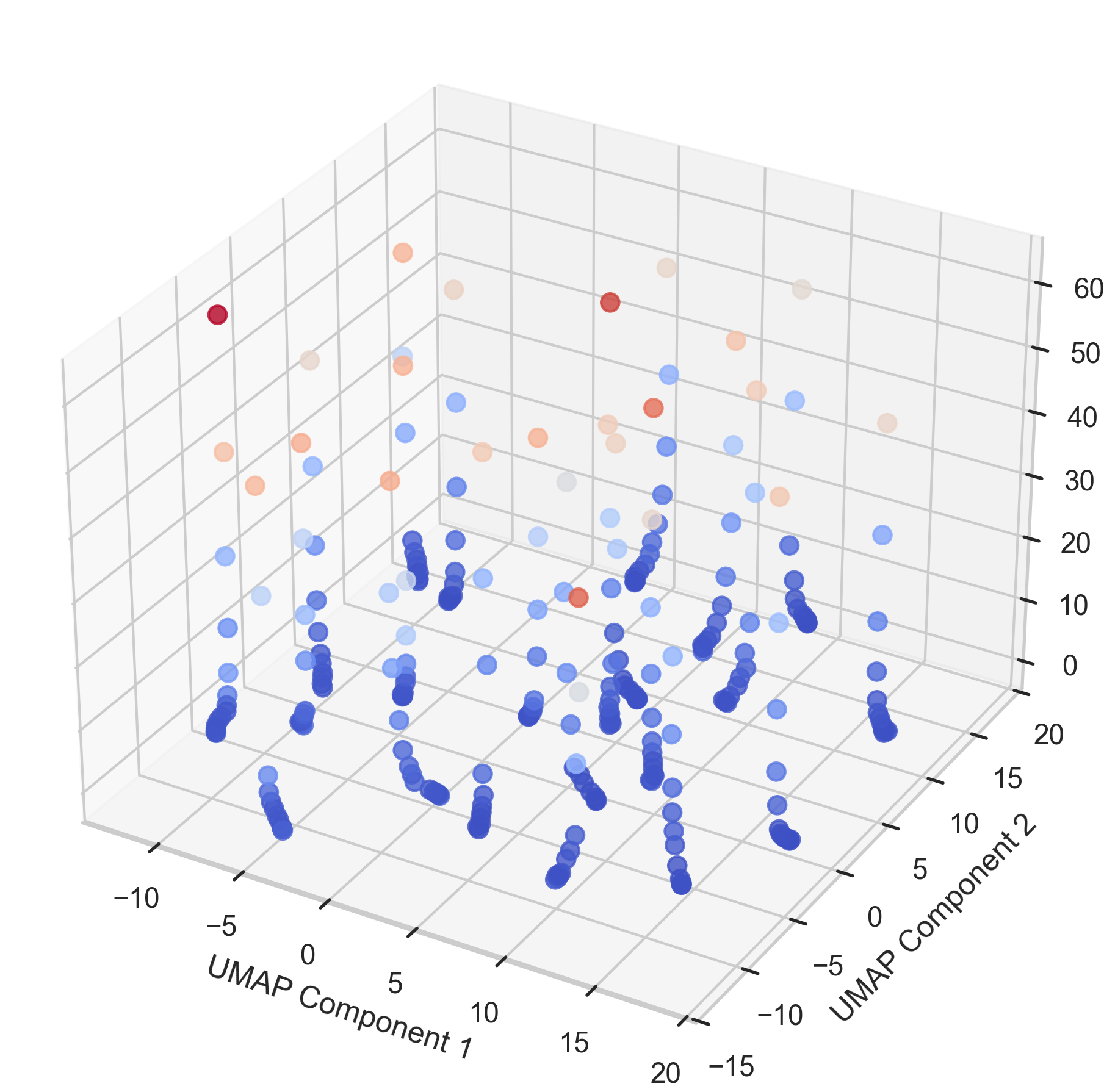}
\caption{IDS}
        \label{fig:umap_yacht_ids}
    \end{subfigure}%
    \hfill
    \begin{subfigure}[b]{0.268\textwidth}
        \centering
        \includegraphics[width=\linewidth]{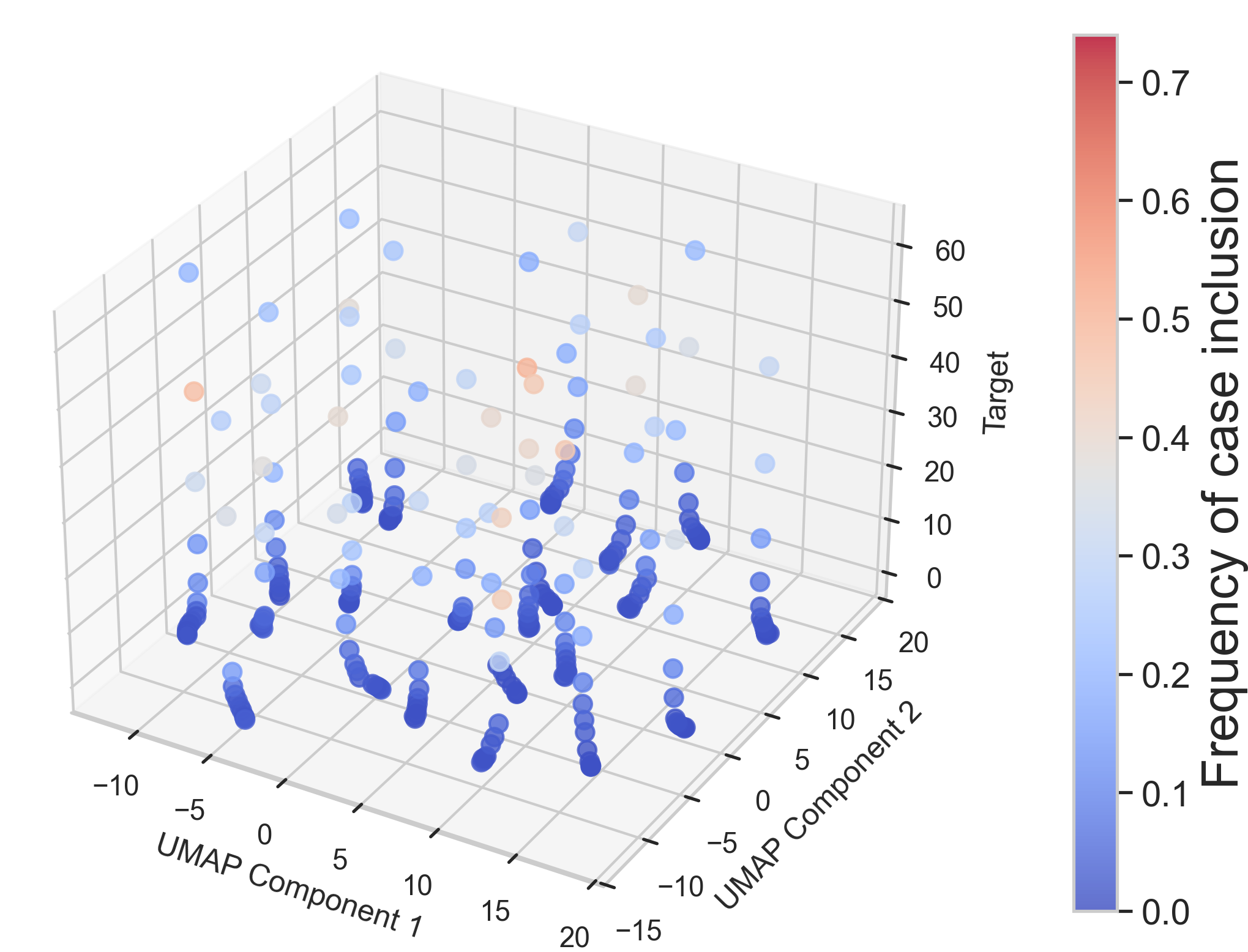}
        \caption{ROIDS}
        \label{fig:umap_yacht_roids}
    \end{subfigure}
    \caption{UMAP Visualization of the \texttt{yacht} problem. The color scale indicates how often each case is selected under IDS and ROIDS, respectively. Potential outliers are frequently included by IDS, which explains its low performance. ROIDS includes those cases less frequently leading to a significantly better performance compared to IDS.}
    \label{fig:umap_yacht}
\end{figure}

For \texttt{yacht} (Fig.~\ref{fig:umap_yacht_ids}), we observe that IDS heavily focuses on a few outliers.    ROIDS mostly excludes those cases from its subsets (see Fig.~\ref{fig:umap_yacht_roids}). This probably explains why IDS performs notably worse than RDS for the \texttt{yacht} problem and why ROIDS outperforms IDS significantly here.
The UMAP visualizations of \texttt{concrete}, \texttt{housing}, and \texttt{redwine} are in the supplementary material in Appendix~\ref{appendix:real_world}, Fig.~\ref{fig:umap_concrete}- ~\ref{fig:umap_redwine}, were we observe similar behavior as discussed above.

To sum up, ROIDS significantly outperforms IDS on two problems, where IDS is worse than RDS. Additionally, ROIDS outperforms IDS even if an outlier detection algorithm is applied to the training set beforehand. Overall, ROIDS achieves the best average ranking on the real-world problems, which highlights the robust performance compared to IDS, making it a good choice across a wide range of problems.

\section{Conclusions and Future Work}\label{sec:conclusions}

Recent work found that IDS as a down-sampling method for selection is beneficial for problems in the program synthesis domain~\cite{boldi2024informed} as well as for symbolic regression~\cite{geiger2025performance}.
Our analysis revealed that this is due to the ability of IDS to build a subset of training cases that better represents underlying patterns even if the data is unevenly distributed in the training set.
However, our analysis also revealed that the performance of IDS degrades in the presence of outliers by over-representing these unwanted cases in its subsets, consequently impeding search performance. 

To overcome this limitation, we introduced ROIDS in this paper. ROIDS retains the advantages of IDS and continues to work well on unevenly distributed data points, but, most importantly, performs significantly better in the presence of outliers thanks to its novel outlier-aware down-sampling strategy.  

In our experiments on the synthetic problems, we find that ROIDS shows exactly the desired behavior: the subsets still focus on edge and corner cases, but outliers are less important and no longer interfere with the evolutionary search. This is also reflected in the results. Whenever we added outliers to the synthetic problems, ROIDS outperformed the results achieved with IDS. 

Moreover, the positive aspects of our new approach are also visible when applied to more complex real-world problems, as ROIDS outperforms IDS on over 80\% of the benchmark problems. 

Given that ROIDS consistently achieves the top average rank across the synthetic as well as the real-world benchmark problems, we recommend researches and practitioners to adopt ROIDS as their preferred down-sampling method for selection in GP in the symbolic regression domain. 
 
In future work, we will analyze ROIDS on further benchmark problems and develop an adaptive version of ROIDS that automatically adapts the outlier sensitivity rate.

\bibliographystyle{ACM-Reference-Format}
\bibliography{main}

\clearpage              
\appendix

\section{Analysis Friedman Problems}\label{appendix:friedman}

\begin{figure}[H]
    \centering
    \includegraphics[width=1.0\linewidth]{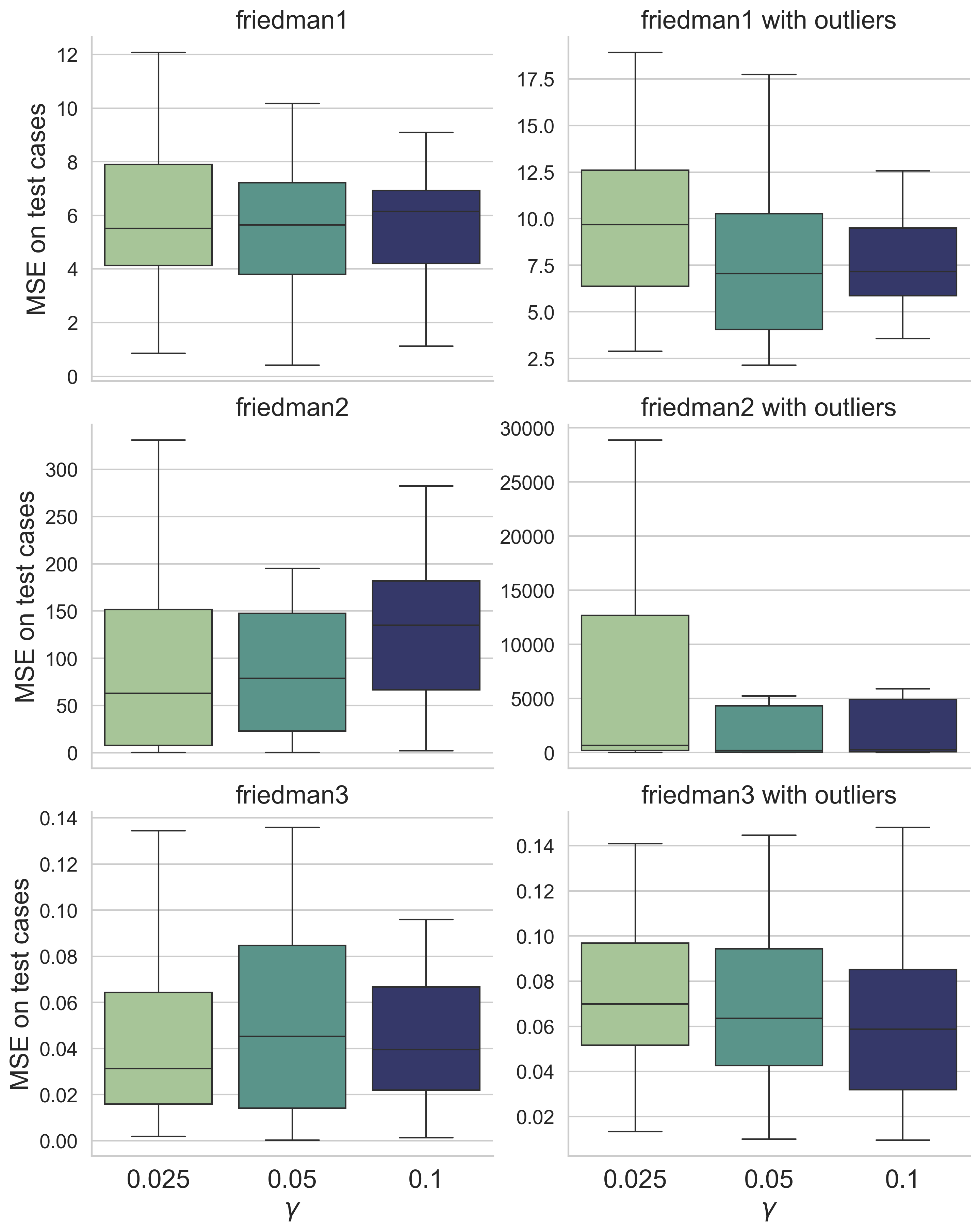}
    \caption{Analysis of the parameter $\gamma$ of ROIDS for all variants of the Friedman problems. The performance over 30 runs is shown.  For better readability, outliers are not plotted.}
    \label{fig:friedman_gammas}
\end{figure}

Fig.~\ref{fig:friedman_gammas} plots the performance of ROIDS for different outlier sensitivity rates $\gamma \in \{0.025, 0.05, 0.1\}$. Results are for all variants of the Friedman problems. For the Friedman problems without outliers, lower values of $\gamma$ lead to higher performance. This is as expected, as these problems only contain cases that represent the underlying true pattern and should not be removed. However, in real-world settings this is rarely the case. 

For Friedman problems containing outliers, a $\gamma$ greater than the percentage of added outliers (here $5\%$) leads to better results. This shows that it is important to set $\gamma$ large enough to remove all existing outliers. A value of $\gamma=0.05$ performs reasonably well across all problem variants.  

\begin{figure}[h]
    \centering
    \begin{subfigure}[b]{0.208\textwidth}
        \centering
        \includegraphics[width=\linewidth]{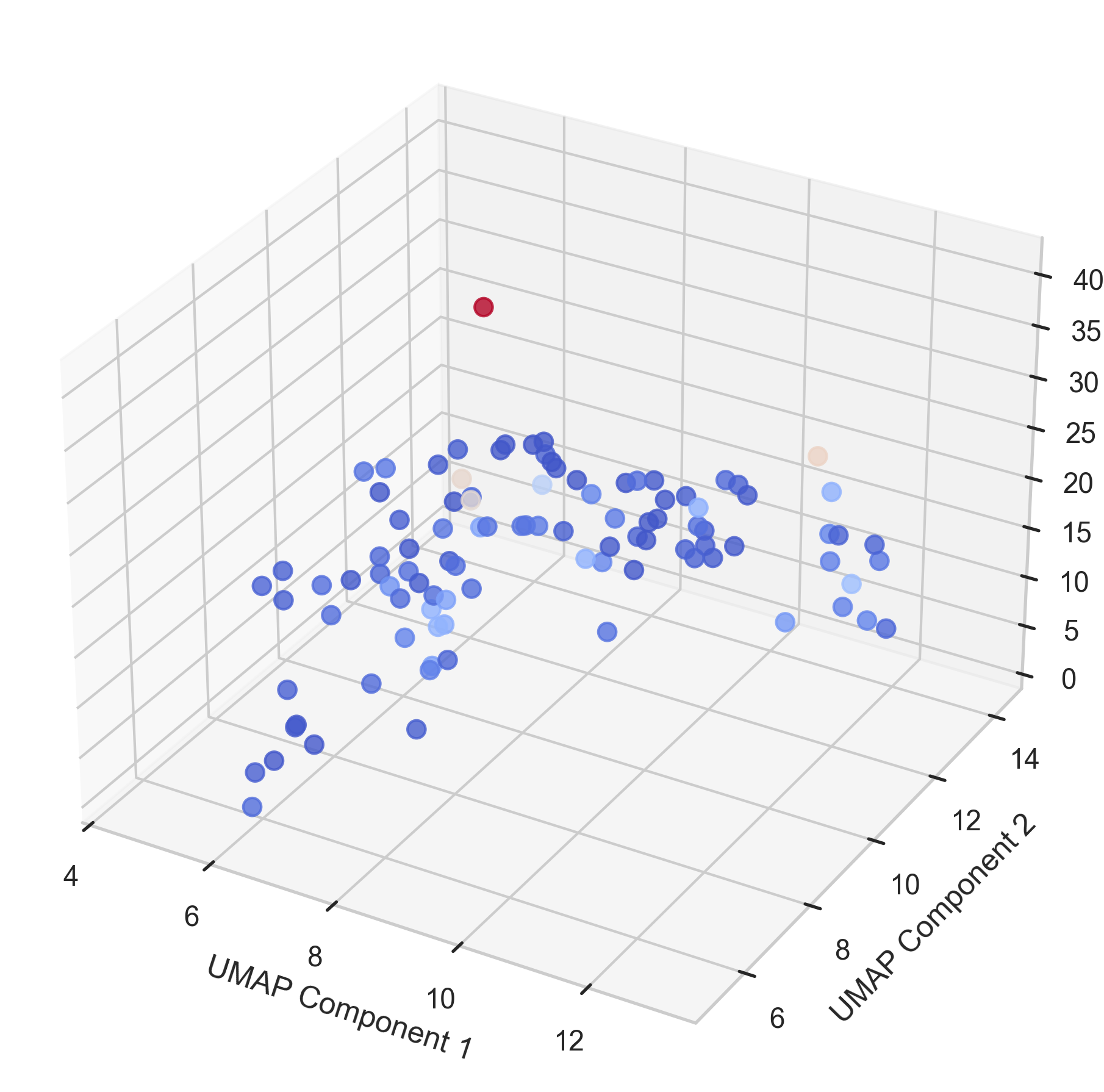}
\caption{IDS}
        \label{fig:umap_friedman1_ids}
    \end{subfigure}%
    \hfill
    \begin{subfigure}[b]{0.268\textwidth}
        \centering
        \includegraphics[width=\linewidth]{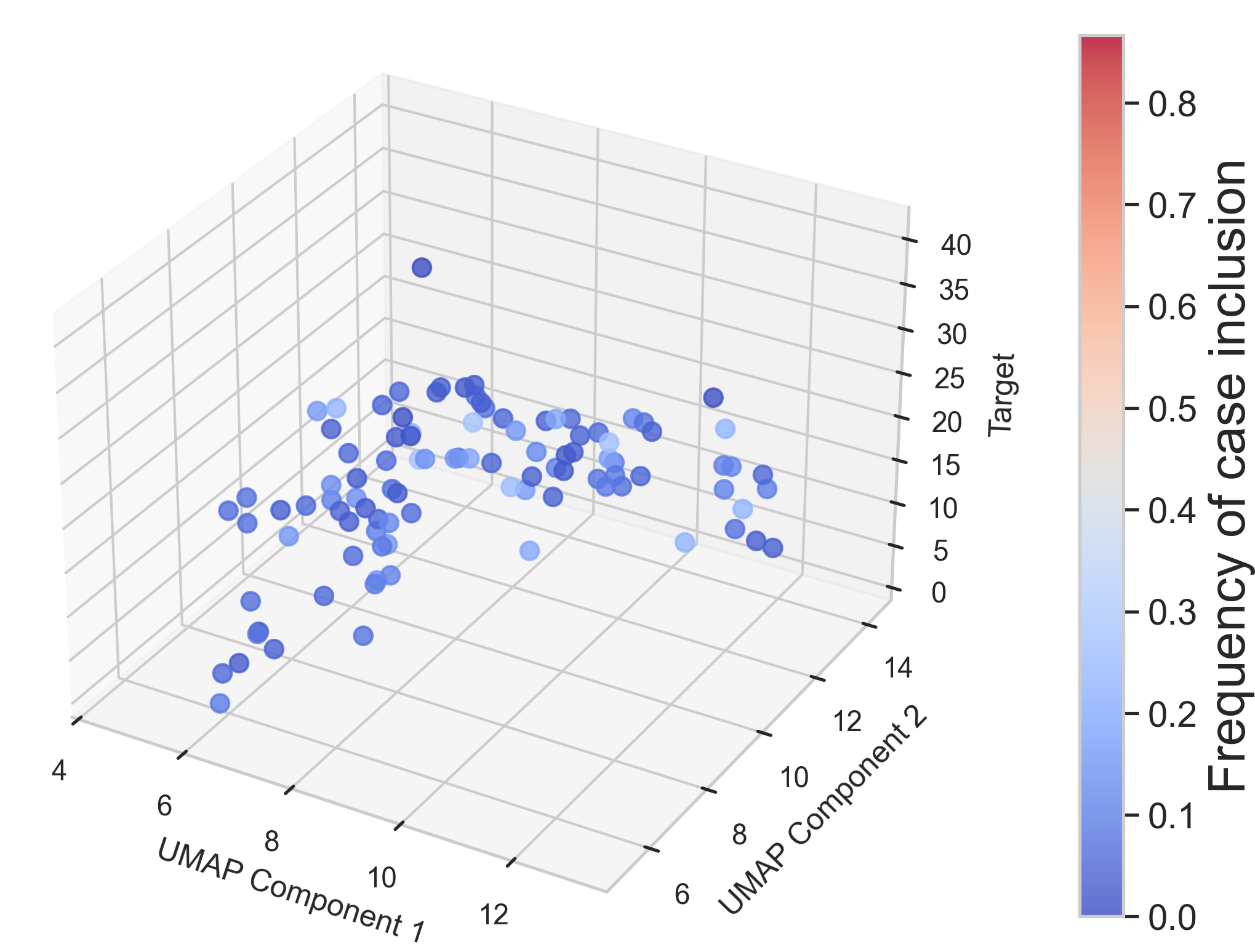}
        \caption{ROIDS}
        \label{fig:umap_friedman1_roids}
    \end{subfigure}
    \caption{UMAP Visualization of the \texttt{friedman1} problem with outliers. The color scale shows how often each case is selected under IDS and ROIDS, respectively.}
    \label{fig:umap_friedman1}
\end{figure}

\begin{figure}[h]
    \centering
    \begin{subfigure}[b]{0.208\textwidth}
        \centering
        \includegraphics[width=\linewidth]{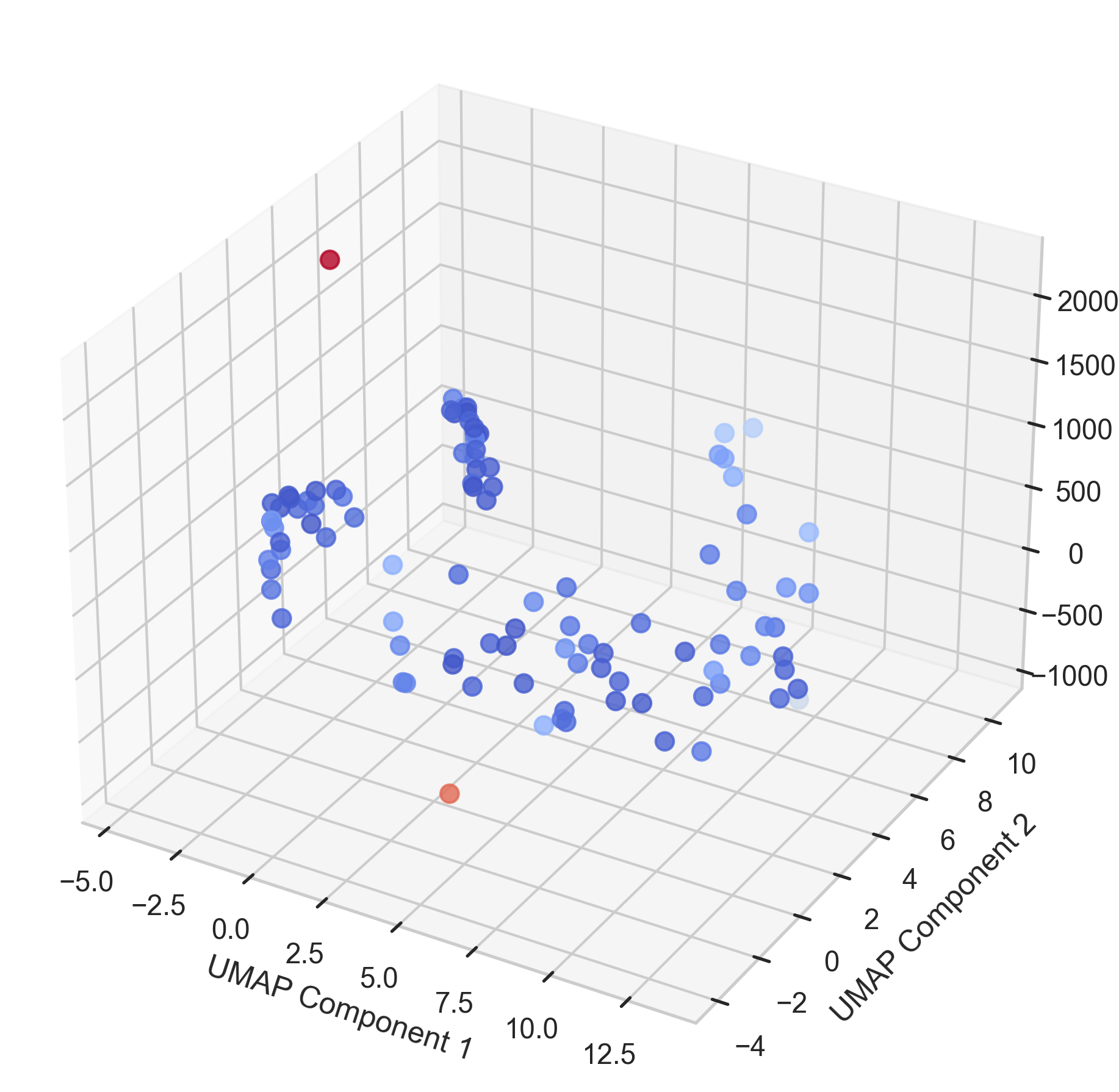}
\caption{IDS}
        \label{fig:umap_friedman2_ids}
    \end{subfigure}%
    \hfill
    \begin{subfigure}[b]{0.268\textwidth}
        \centering
        \includegraphics[width=\linewidth]{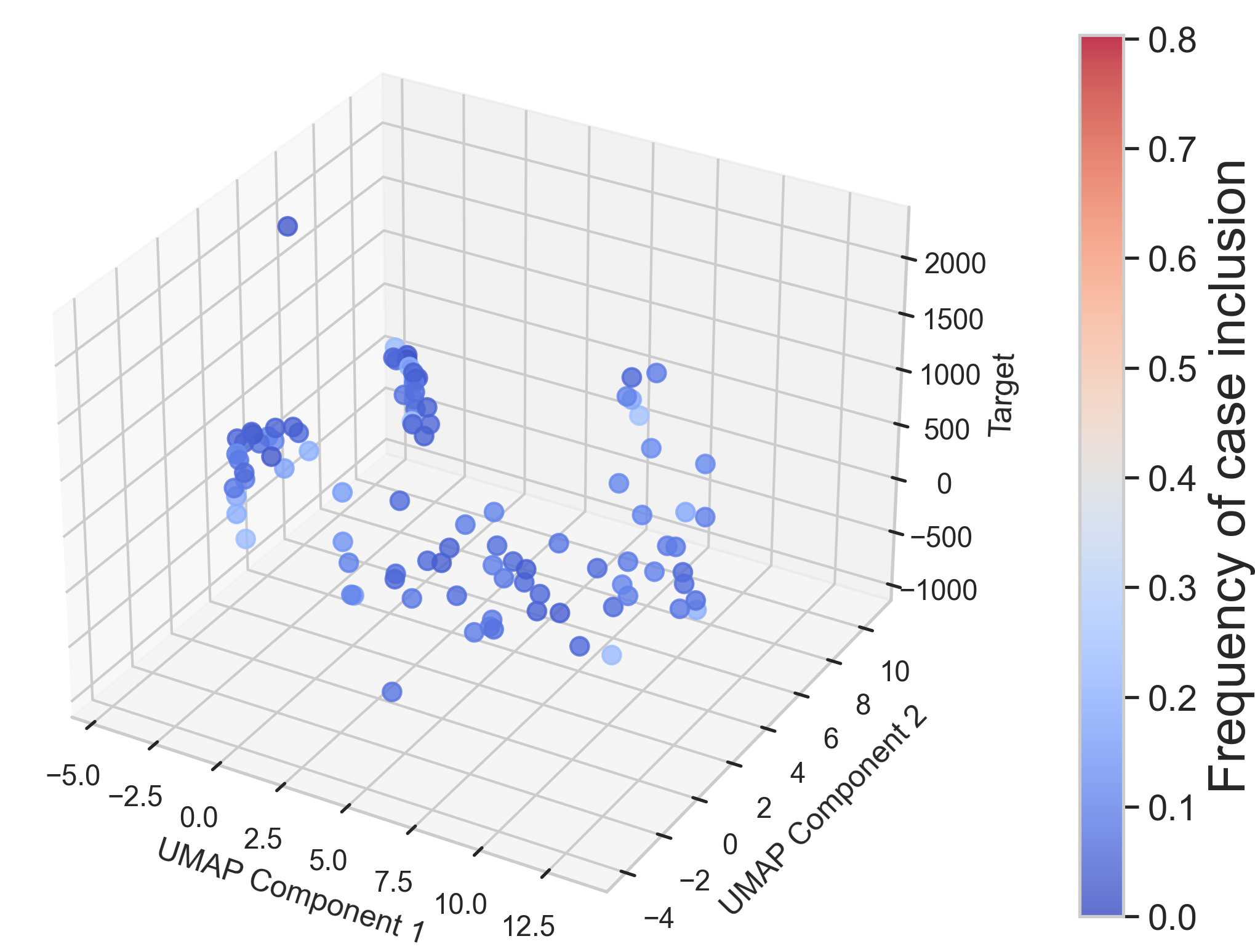}
        \caption{ROIDS}
        \label{fig:umap_friedman2_roids}
    \end{subfigure}
    \caption{UMAP Visualization of the \texttt{friedman2} problem with outliers. The color scale shows how often each case is selected under IDS and ROIDS, respectively.}
    \label{fig:umap_friedman2}
\end{figure}

\begin{figure}[h]
    \centering
    \begin{subfigure}[b]{0.208\textwidth}
        \centering
        \includegraphics[width=\linewidth]{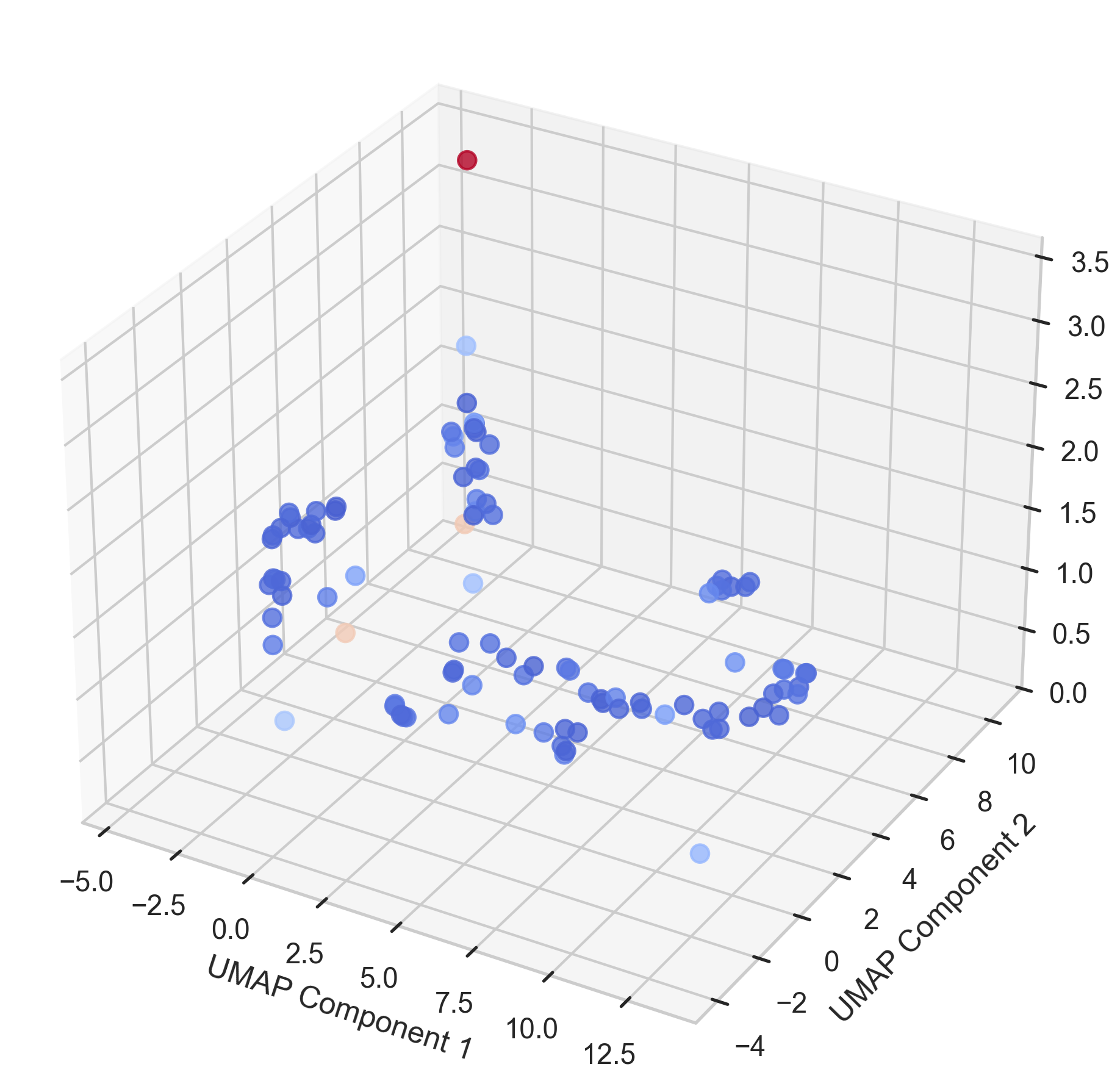}
\caption{IDS}
        \label{fig:umap_friedman3_ids}
    \end{subfigure}%
    \hfill
    \begin{subfigure}[b]{0.268\textwidth}
        \centering
        \includegraphics[width=\linewidth]{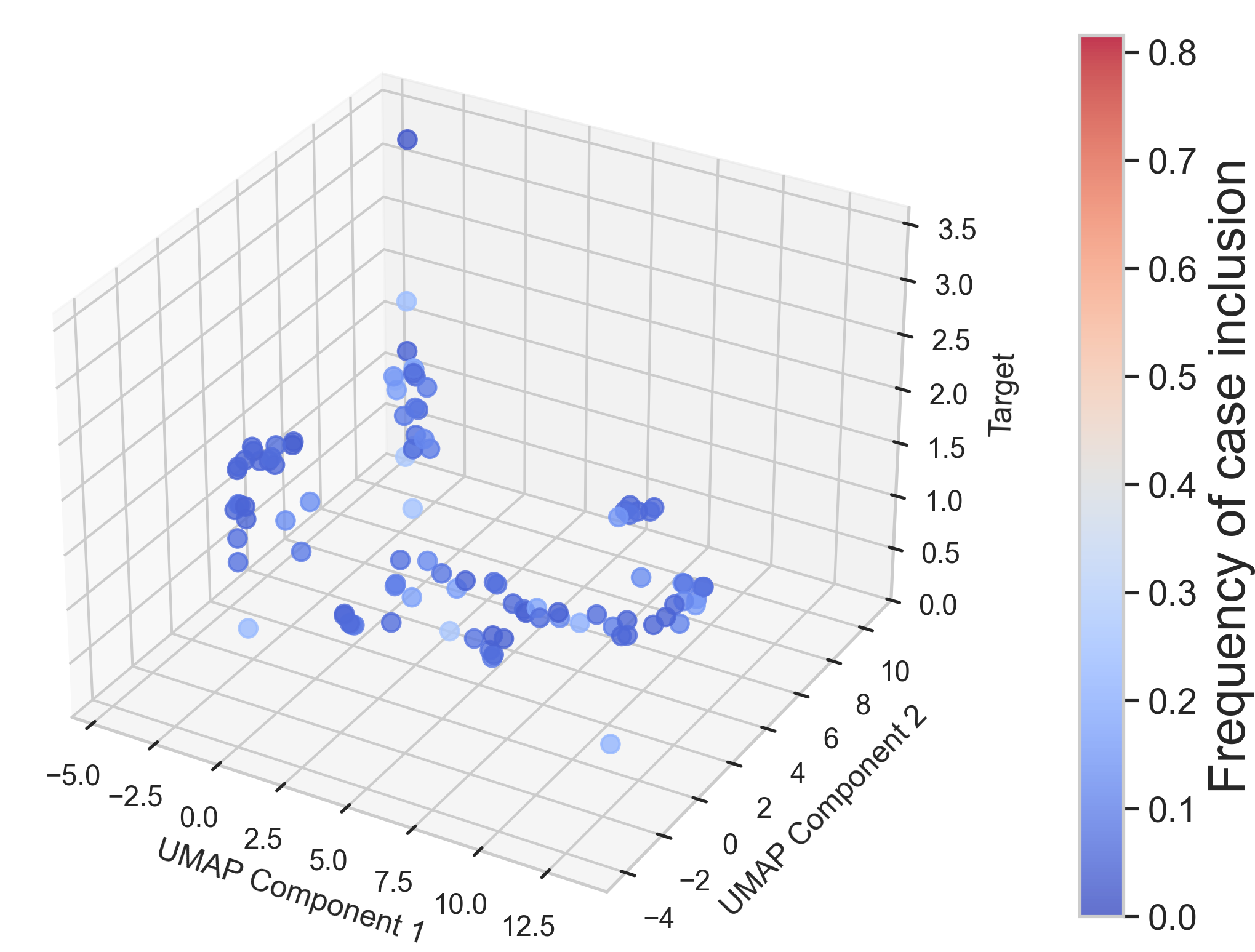}
        \caption{ROIDS}
        \label{fig:umap_friedman3_roids}
    \end{subfigure}
    \caption{UMAP Visualization of the \texttt{friedman3} problem with outliers. The color scale shows how often each case is selected under IDS and ROIDS, respectively.}
    \label{fig:umap_friedman3}
\end{figure}

Figures~\ref{fig:umap_friedman1}-\ref{fig:umap_friedman3} show UMAP visualizations of the \texttt{friedman1}, \\ \texttt{friedman2}, and \texttt{friedman3} problems with outliers. The color encodes the frequency of case inclusion by IDS and ROIDS. We observe that outliers are frequently included in the subsets created by IDS (Figs.~\ref{fig:umap_friedman1_ids}, ~\ref{fig:umap_friedman2_ids}, and ~\ref{fig:umap_friedman3_ids}). In contrast, ROIDS often excludes those cases from its subsets and focuses on including cases that represent the true underlying pattern of the problem (Figs.~\ref{fig:umap_friedman1_roids}, ~\ref{fig:umap_friedman2_roids}, and~\ref{fig:umap_friedman3_roids}).

\section{Analysis Real-world Problems}\label{appendix:real_world}

\begin{figure}[h]
    \centering
    \includegraphics[width=1.0\linewidth]{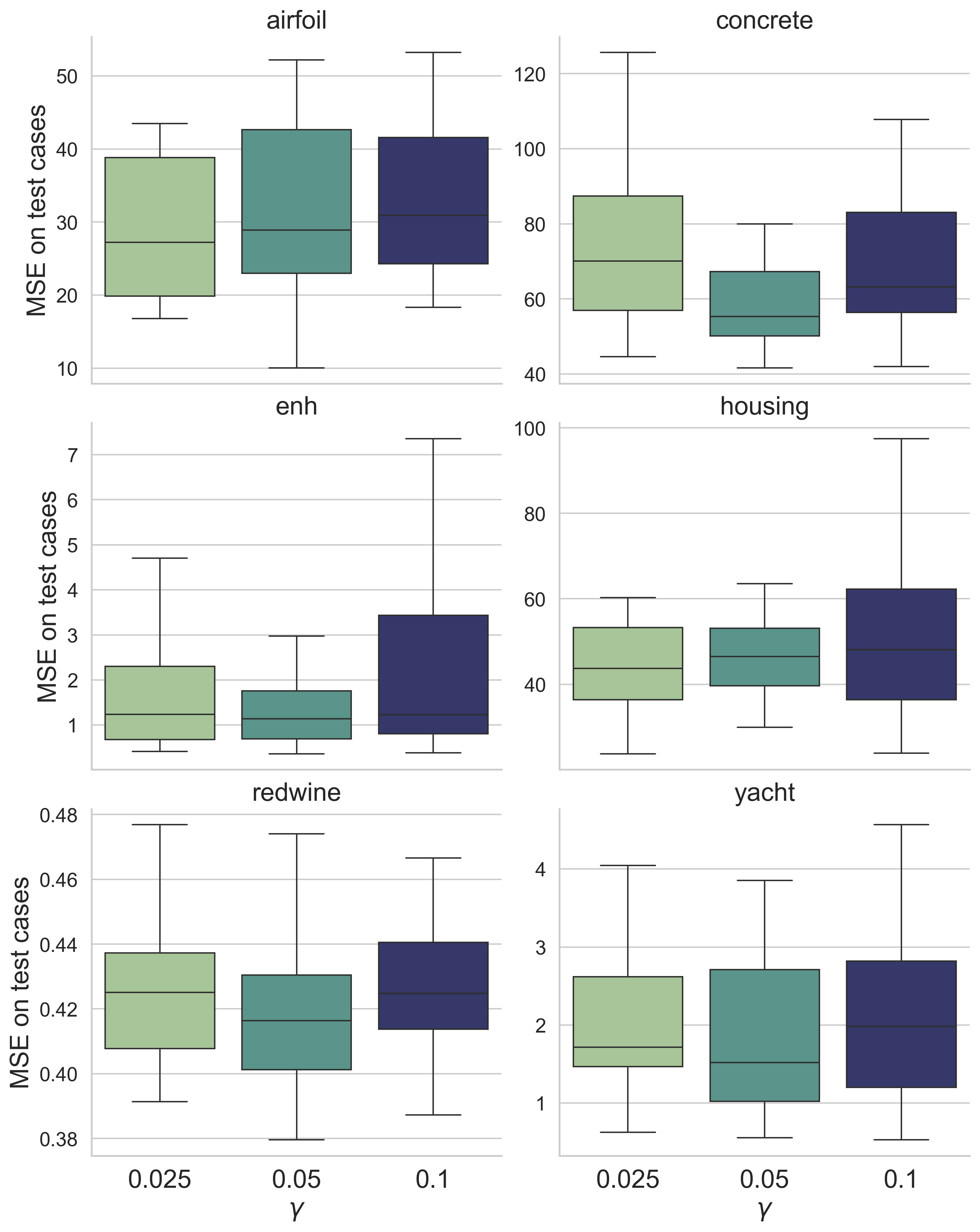}
    \caption{Analysis of the parameter $\gamma$ of ROIDS for all real‑world problems. The performance over 30 runs is shown.  For better readability, outliers are not plotted.}
    \label{fig:realworld_gammas}
\end{figure}

Figure~\ref{fig:realworld_gammas} plots the performance of ROIDS for $\gamma \in \{0.025, 0.05, 0.1\}$ for all real-world problems. For the \texttt{concrete}, \texttt{enh}, \texttt{redwine}, and \texttt{yacht} problem, $\gamma = 0.05$ achieves the lowest MSE. Overall, $\gamma$ is robust over a reasonable range for real-world problems.

\begin{figure}[htbp]
    \centering
    \begin{subfigure}[b]{0.208\textwidth}
        \centering
        \includegraphics[width=\linewidth]{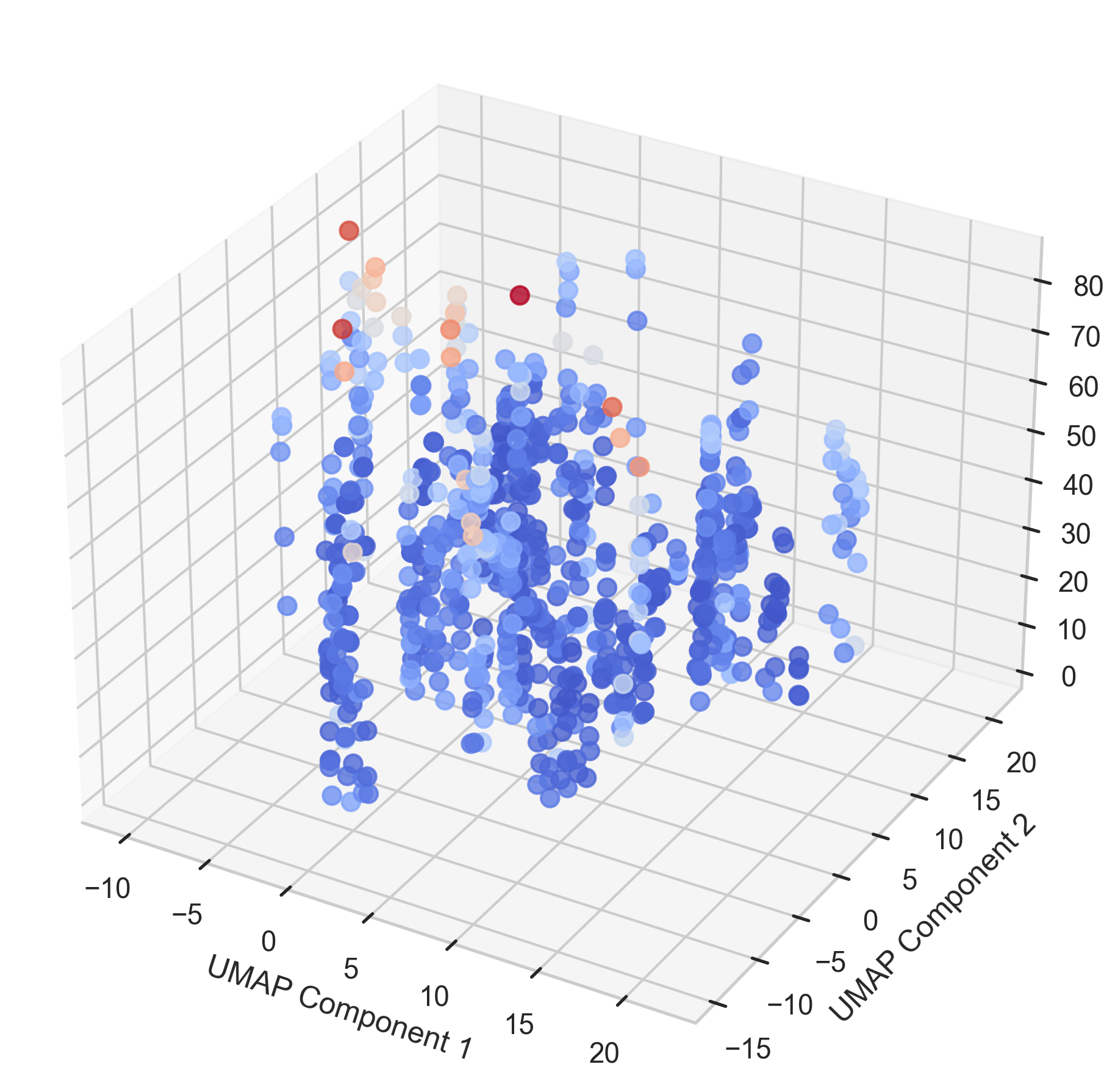}
\caption{IDS}
        \label{fig:umap_concrete_ids}
    \end{subfigure}%
    \hfill
    \begin{subfigure}[b]{0.268\textwidth}
        \centering
        \includegraphics[width=\linewidth]{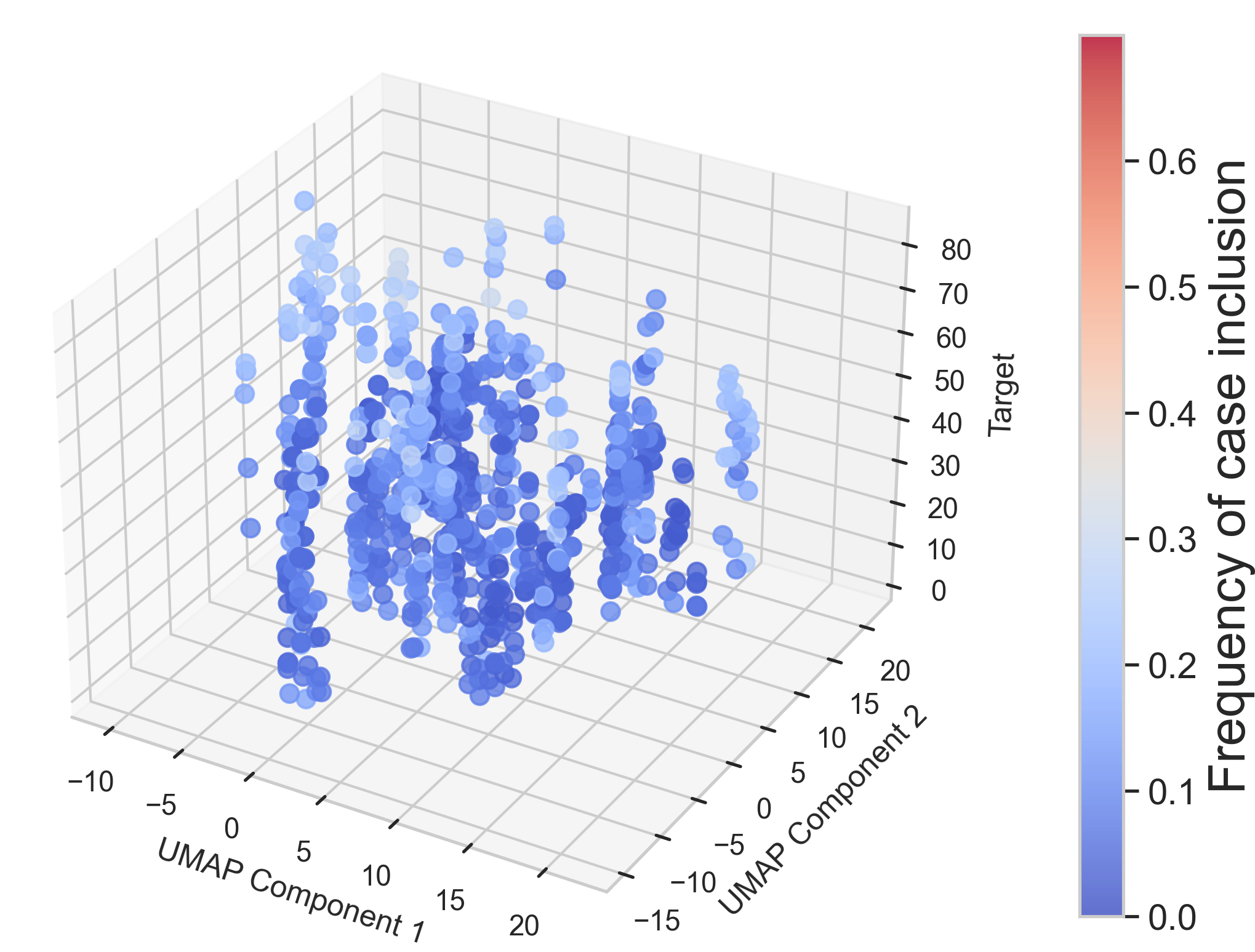}
        \caption{ROIDS}
        \label{fig:umap_concrete_roids}
    \end{subfigure}
    \caption{UMAP Visualization of the \texttt{concrete} problem. The color scale shows how often each case is selected under IDS and ROIDS, respectively.}
    \label{fig:umap_concrete}
\end{figure}

Further, we visualize the real-world problems using UMAPs. For the \texttt{concrete} problem (Fig.~\ref{fig:umap_concrete}), all training cases lie in one dense region. Here, all down-sampling methods perform equally well. IDS includes some of the edge cases more frequently (see Fig.~\ref{fig:umap_concrete_ids}), however, this does not impact performance.

\begin{figure}[htbp]
    \centering
    \begin{subfigure}[b]{0.208\textwidth}
        \centering
        \includegraphics[width=\linewidth]{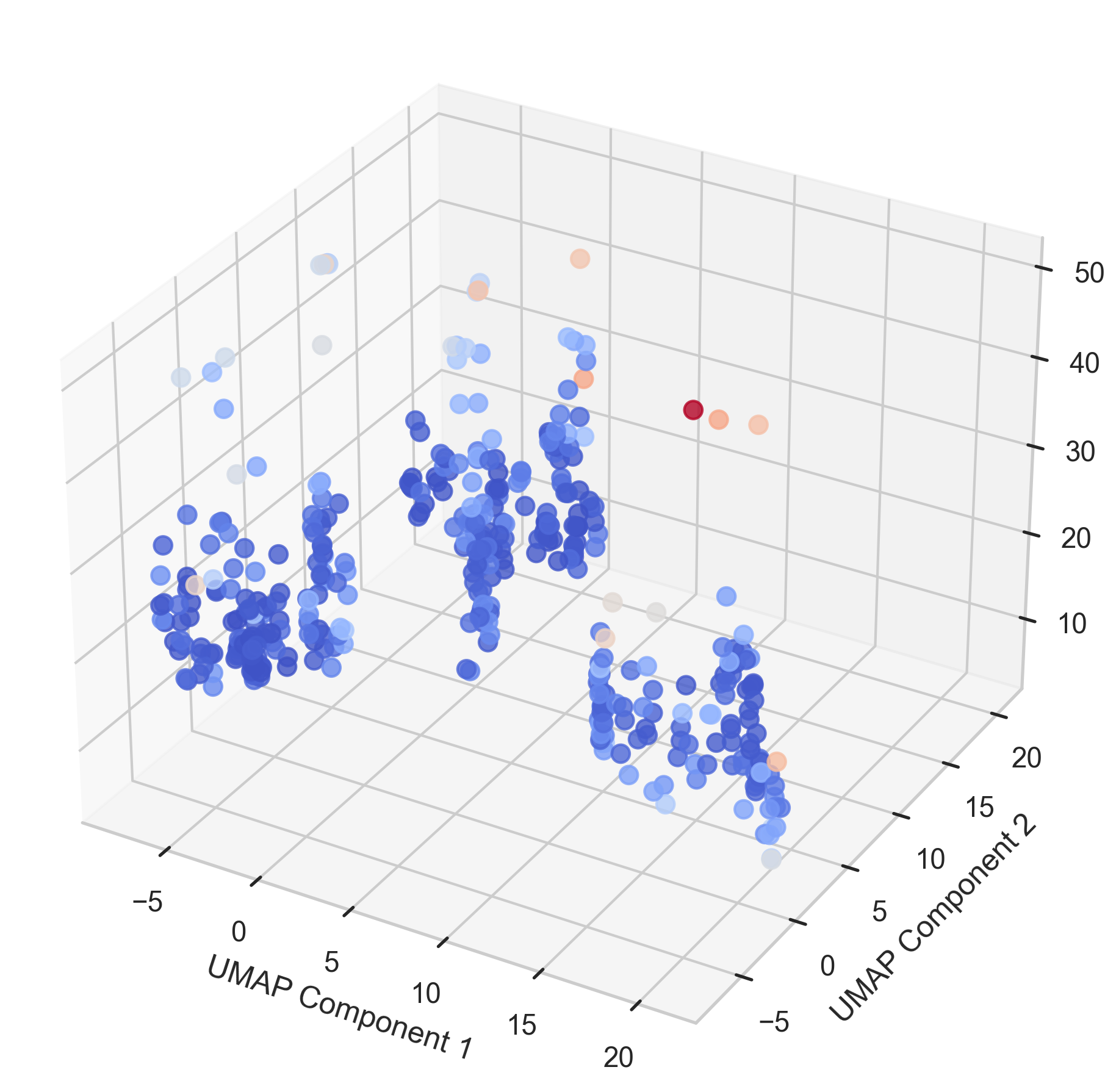}
\caption{IDS}
        \label{fig:umap_housing_ids}
    \end{subfigure}%
    \hfill
    \begin{subfigure}[b]{0.268\textwidth}
        \centering
        \includegraphics[width=\linewidth]{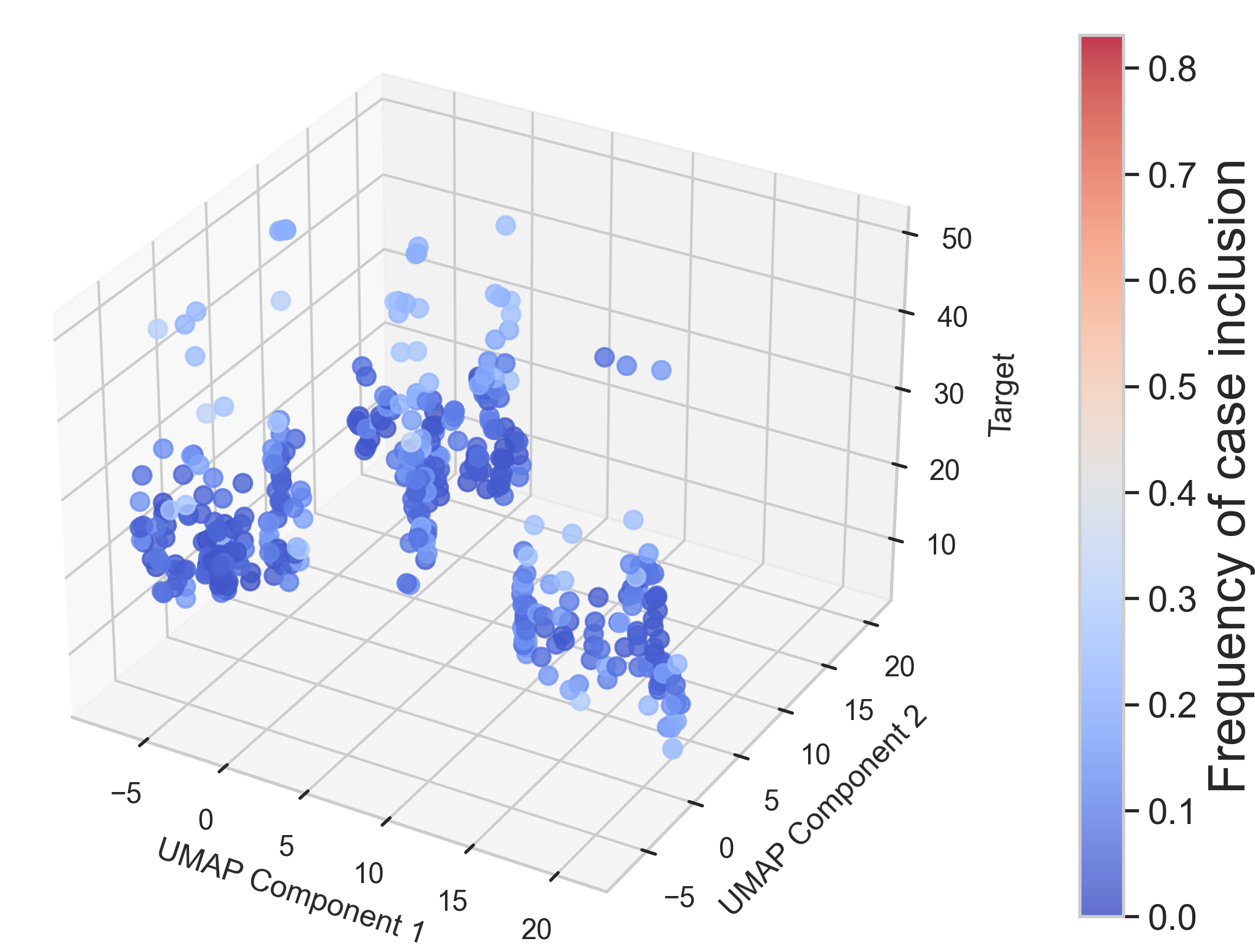}
        \caption{ROIDS}
        \label{fig:umap_housing_roids}
    \end{subfigure}
    \caption{UMAP Visualization of the \texttt{housing} problem. The color scale shows how often each case is selected under IDS and ROIDS, respectively.}
    \label{fig:umap_housing}
\end{figure}

For the \texttt{housing} problem (Fig.~\ref{fig:umap_housing}), we observe a few visual outlier cases. Those are frequently included in the subsets constructed by IDS (see Fig.~\ref{fig:umap_housing_ids}), which explains why IDS performs poorly for this problem. In contrast, ROIDS excludes those cases from its subsets most of the time (see Fig.~\ref{fig:umap_housing_roids}) leading to a better performance compared to IDS. 

\begin{figure}[htbp]
    \centering
    \begin{subfigure}[b]{0.208\textwidth}
        \centering
        \includegraphics[width=\linewidth]{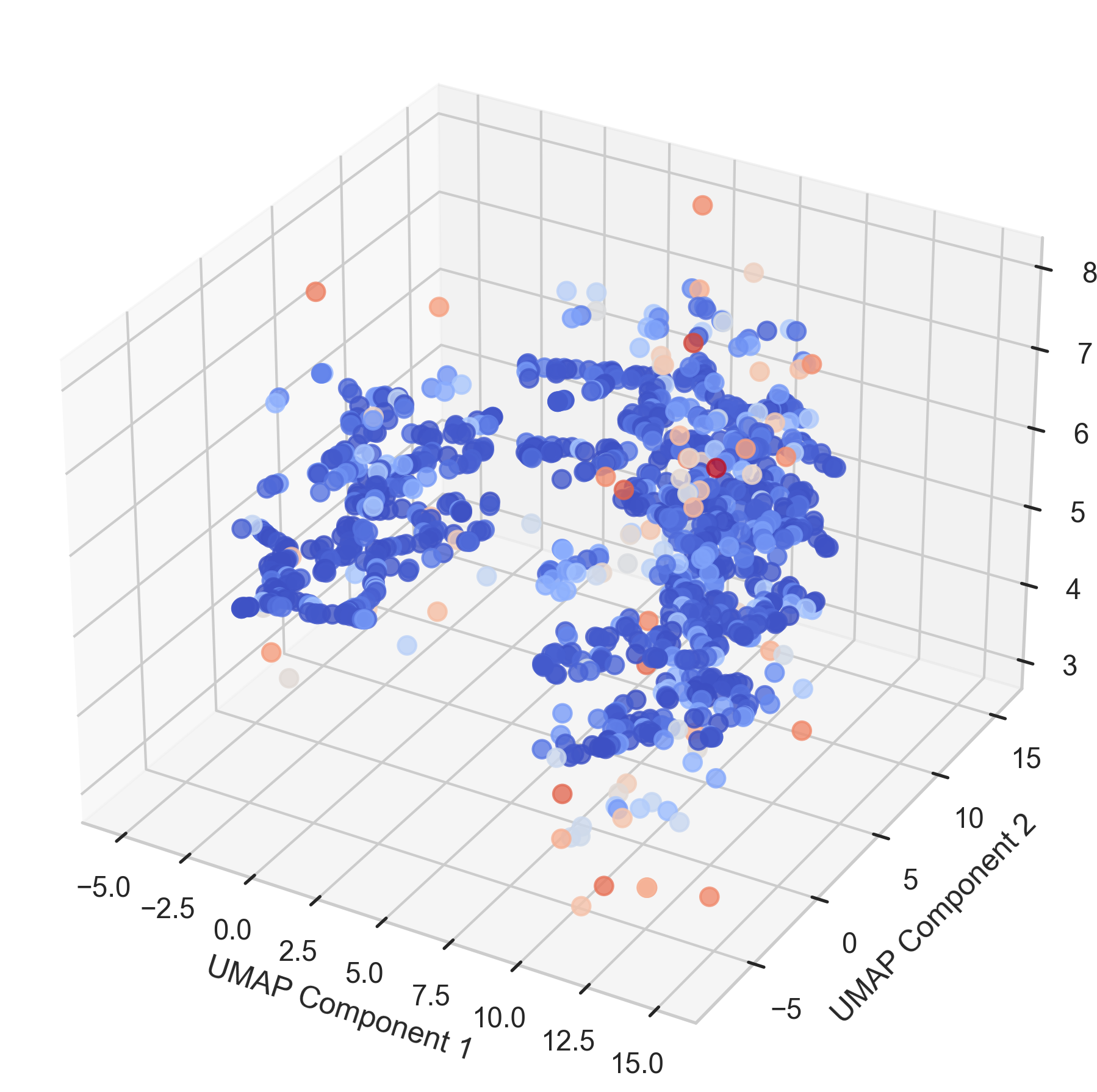}
\caption{IDS}
        \label{fig:umap_redwine_ids}
    \end{subfigure}%
    \hfill
    \begin{subfigure}[b]{0.268\textwidth}
        \centering
        \includegraphics[width=\linewidth]{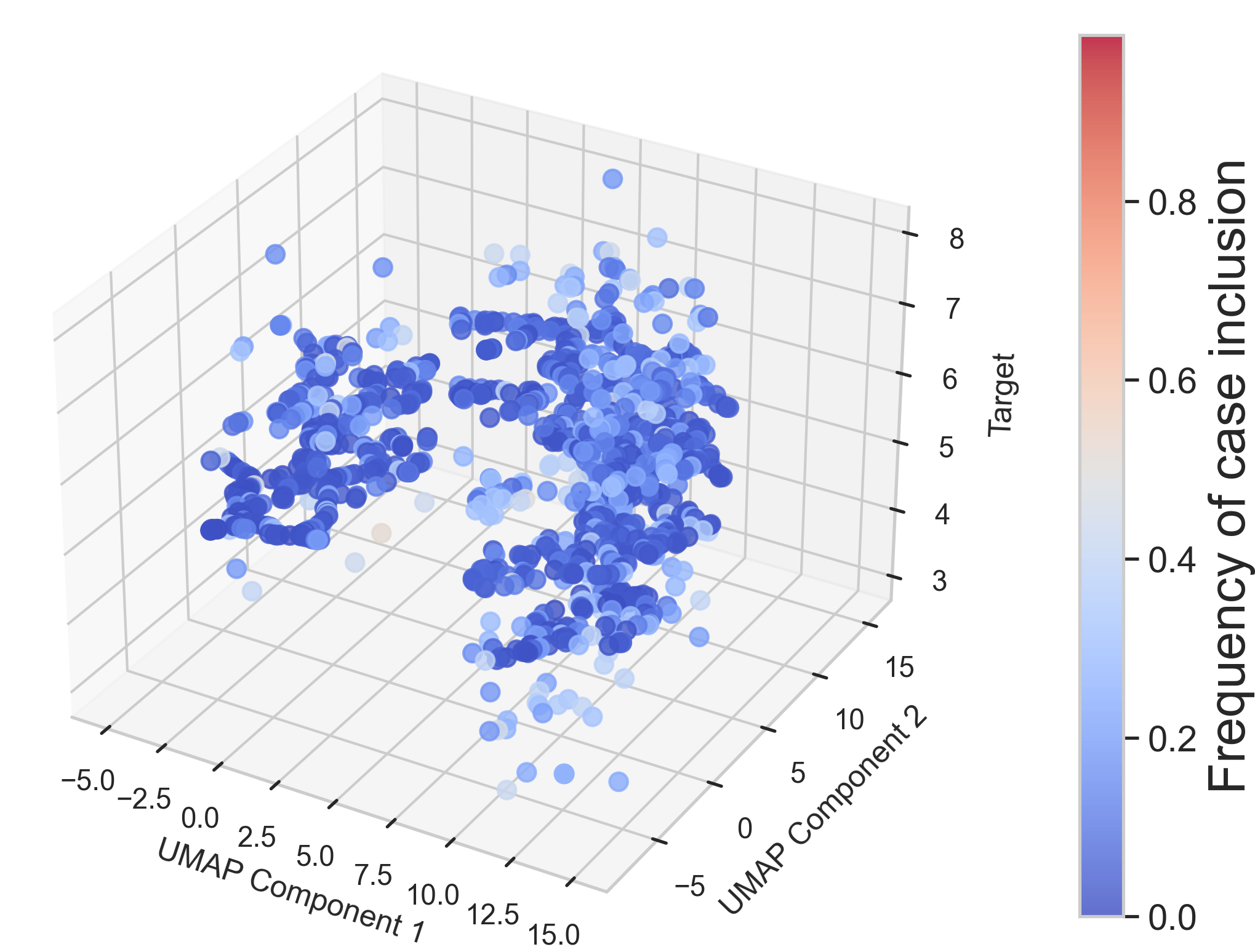}
        \caption{ROIDS}
        \label{fig:umap_redwine_roids}
    \end{subfigure}
    \caption{UMAP Visualization of the \texttt{redwine} problem. The color scale shows how often each case is selected under IDS and ROIDS, respectively.}
    \label{fig:umap_redwine}
\end{figure}

Figure~\ref{fig:umap_redwine} visualizes the distribution of the data points for the \texttt{redwine} problem. Here, IDS includes edge cases and potential outliers more frequently (see Fig.~\ref{fig:umap_redwine_ids}). ROIDS performs significantly better because it includes potential outliers less frequently in its subsets (see Fig.~\ref{fig:umap_redwine_roids}).


\end{document}